\documentclass[10pt,twocolumn,letterpaper]{article}

\usepackage[pagenumbers]{cvpr} 
\usepackage{amsthm}
\newtheorem*{customtheorem}{Theorem}
\usepackage{tikz}
\usepackage{subcaption}
\usepackage{rotating}
\usepackage{placeins}
\usepackage{float}
\usepackage{graphicx}
\definecolor{cvprblue}{rgb}{0.21,0.49,0.74}
\usepackage[pagebackref,breaklinks,colorlinks,allcolors=cvprblue]{hyperref}

\def\paperID{3523} 
\def\confName{CVPR}
\def\confYear{2026}

\title{Masked Auto-Regressive Variational Acceleration:\\
Fast Inference Makes Practical Reinforcement Learning}

\author{Yuxuan Gu$^{1}$\textsuperscript{\dag}, \:Weimin Bai$^1$\textsuperscript{\dag}, \:Yifei Wang$^2$, \:Weijian Luo$^3$, \:He Sun$^1$\textsuperscript{\ddag} \\[2pt]
1. Peking University \quad
2. Rice University \quad
3. hi-lab, Xiaohongshu Inc \\ [2pt]
    \vspace{0.05em}  
}
\vspace{-0.15em}

\begin{document}
\maketitle

\begingroup
\renewcommand\thefootnote{} 
%
%
\footnotetext{\textsuperscript{\dag}Equal Contributions. \quad \textsuperscript{\ddag}Corresponding author.}
\endgroup

\begin{abstract}

Masked auto-regressive diffusion models (MAR) benefit from the expressive modeling ability of diffusion models and the flexibility of masked auto-regressive ordering. However, vanilla MAR suffers from slow inference due to its hierarchical inference mechanism: an outer AR unmasking loop and an inner diffusion denoising chain. Such decoupled structure not only harm the generation efficiency but also hinder the practical use of MAR for reinforcement learning (RL), an increasingly critical paradigm for generative model post-training.
To address this fundamental issue, we introduce MARVAL (Masked Auto-regressive Variational Acceleration), a distillation-based framework that compresses the diffusion chain into a single AR generation step while preserving the flexible auto-regressive unmasking order. Such a distillation with MARVAL not only yields substantial inference acceleration but, crucially, makes RL post-training with verifiable rewards practical, resulting in scalable yet human-preferred fast generative models. Our contributions are twofold: (1) a novel score-based variational objective for distilling masked auto-regressive diffusion models into a single generation step without sacrificing sample quality; and (2) an efficient RL framework for masked auto-regressive models via MARVAL-RL. On ImageNet 256×256, MARVAL-Huge achieves an FID of \textbf{2.00} with more than \textbf{30 times speedup} compared with MAR-diffusion, and MARVAL-RL yields consistent improvements in CLIP and image-reward scores on ImageNet datasets with entity names. In conclusion, MARVAL demonstrates the first practical path to distillation and RL of masked auto-regressive diffusion models, enabling fast sampling and better preference alignments. Project page: \url{https://ai4scientificimaging.org/MARVAL-RL/}.

\end{abstract}    
\begin{figure*}[t]
  \centering
  \includegraphics[width=1 \linewidth]{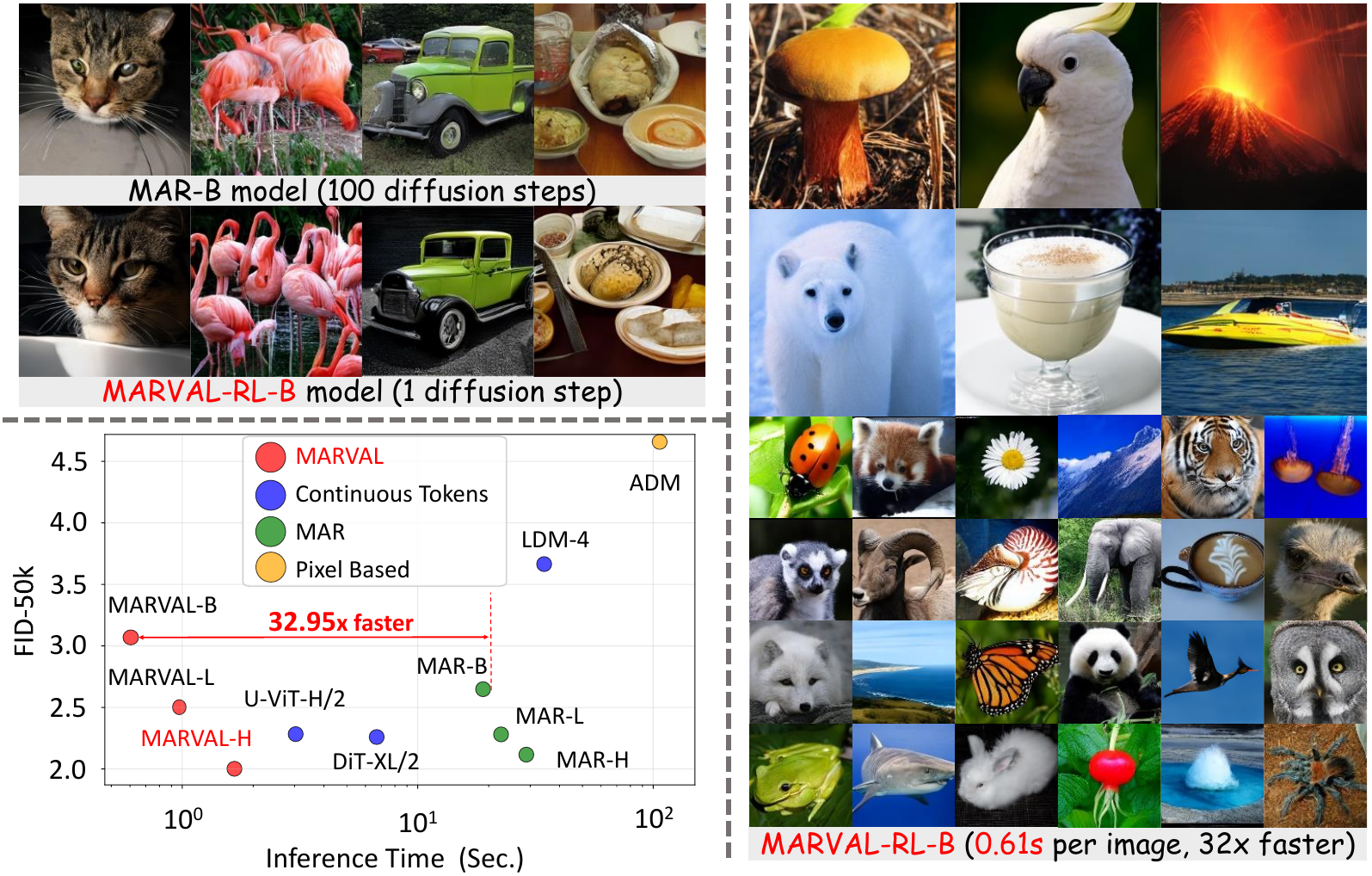}
  \caption{Performance and qualitative results of MARVAL-RL. \textbf{Top-left:} Comparison of image quality between the MAR-B model (100 diffusion steps) and the MARVAL-RL-B model (1 diffusion step). MARVAL-RL significantly surpasses the MAR-B model in semantic quality, fidelity, and clarity. \textbf{Bottom-left:} Comparing FID-50k (y-axis) and inference time generating one image(x-axis) against other state-of-the-art methods. The MARVAL series (red dots) demonstrates superior performance, achieving low FID scores with significantly faster inference speeds (e.g., MARVAL-H achieves a FID of 2.00, and MARVAL-B is 32.95x faster than MAR-B). \textbf{Right: }A collection of diverse, high-quality images generated by MARVAL-RL-B model, showcasing its strong generative capabilities at an average speed of only 0.61 seconds per image.}
  \label{fig:teaser}
\end{figure*}

\section{Introduction}

Auto-regressive (AR)~\cite{van2016pixel, reed2017few, van2016conditional,reed2017parallel, tian2024visual} and diffusion-based~\cite{dhariwal2021diffusion, song2020score}  generative models exhibit complementary strengths and weaknesses. Diffusion models excel at high-fidelity synthesis and mode coverage, but require long denoising trajectories with hundreds of iterative updates~\cite{ho2020denoising, salimans2016improved,song2020denoising}, making inference and downstream Reinforcement Learning (RL) post-training computationally expensive. In contrast, AR models generate sequentially in discrete space, offering natural support for semantic controllability and compatibility with RL and language-based alignment frameworks~\cite{esser2021taming,ramesh2021zero,luong2015effective}, but they often struggle to match the perceptual quality of state-of-the-art diffusion models at high resolutions. Motivated by their complementary properties, recent works have combined the two paradigms—forming AR+Diffusion~\cite{hoogeboom2021autoregressive, gu2022vector, li2024autoregressive} architectures that use AR priors to capture global structure and diffusion decoders to restore fine-grained details. The Masked Auto-regressive (MAR) model~\cite{li2024autoregressive} represents a prominent example of this hybrid design.

However, MAR models face two fundamental limitations. First, their nested architecture—where each AR iteration contains a full diffusion unmasking process—results in extremely slow inference. Second, this inefficiency significantly hinders practices to perform RL-based post-training, which is essential for scalable generative models to be aligned with human preferences\citep{christiano2017deep, ouyang2022training, stiennon2020learning}. The above two issues encourage us to answer a fundamental research question: \textbf{\emph{Can we improve the efficiency of MAR-like models while making RL post-training practical and efficient?}}


To this end, we propose \textbf{\emph{MARVAL}} (\textbf{\emph{M}}asked \textbf{\emph{A}}uto-\textbf{\emph{R}}egressive \textbf{\emph{V}}ariational \textbf{\emph{A}}cce\textbf{\emph{l}}eration), which introduces a novel score-based variational distillation principle that compresses the costly diffusion chain into a single masked-AR generation step, preserving the original AR ordering while achieving more than 30× speed-up. Building upon this acceleration, we further develop MARVAL-RL, an RL fine-tuning algorithm that incorporates reward feedback to enhance fidelity and preference alignment.

Empirically, MARVAL-H achieves competitive ImageNet~\cite{deng2009imagenet} $256\times256$ generation quality (FID = 2.00), which slightly outperforms MAR-H (FID = 2.06) while being 20× faster. When post-trained with RL using human preference rewards, MARVAL-RL further improves CLIP and Image Reward scores, demonstrating better alignment with human preferences. As illustrated in Fig.~\ref{fig:teaser}, MARVAL-RL delivers visually sharper and semantically richer generations than the MAR baseline while maintaining real-time inference efficiency ($\approx 0.6$ s per image). Together, these results establish MARVAL as a practical and scalable path toward efficient, RL-ready image generative modeling with a masked auto-regressive generation order.

\section{Related Works}
\label{sec:relatedworks}

\subsection{Diffusion Models}
Diffusion models \cite{ho2020denoising,song2020score,song2020denoising} try to train a neural network to model data distribution by gradually corrupting samples with a Gaussian forward diffusion process. 
From a continuous-time perspective, Song~\etal~\cite{song2020score} generalized diffusion models using stochastic differential equations (SDEs). The forward perturbation process is an Itô SDE:
\begin{equation}
\label{eq:sde}
\mathrm{d} x= f(x_t,t)\,\mathrm{d}t + g(t)\,\mathrm{d}w,
\end{equation}
where $f(x,t)$ controls the drift and $g(t)$ scales the Brownian motion
$w$ at time $t$. Sampling requires solving the corresponding reverse-time SDE:
\begin{equation}
\label{eq:reverse_sde}
\mathrm{d}\boldsymbol{x}_t = \left[{f}(\boldsymbol{x}_t, t) - g(t)^2 \nabla_{\boldsymbol{x}_t}\log p_t(\boldsymbol{x}_t)\right]\mathrm{d}t + g(t)\mathrm{d}\overline{\boldsymbol{w}},
\end{equation}
which explicitly involves the score function $\nabla_{\boldsymbol{x}_t}\log p_t(\boldsymbol{x}_t)$. Since this quantity is unknown, it is approximated by a neural network $\boldsymbol{S}_\theta(\boldsymbol{x}_t,t)$ by minimizing a weighted denoising score matching loss~\cite{song2019generative}:
\begin{equation}
\label{eq:score_loss}
\mathcal{L}(\theta) = \mathbb{E}_{t,\boldsymbol{x}_t} \left[ \lambda(t) \left\| \boldsymbol{S}_\theta(\boldsymbol{x}_t, t) - \nabla_{\boldsymbol{x}_t}\log p_{t}(\boldsymbol{x}_t|\boldsymbol{x}_0) \right\|_2^2 \right],
\end{equation}
where the weighting function $\lambda(t)$ balances different noise levels.
Once trained, the score network can be directly integrated into Eq.~\ref{eq:reverse_sde} to simulate reverse SDEs~\cite{song2020score,karras2022elucidating,xue2023sa} or ODEs~\cite{song2020denoising,lu2022dpm}, thereby generating high-quality samples.


\subsection{Masked Auto-regressive Model}
The most common practice of traditional diffusion models tries to model the entire image simultaneously in some latent space \cite{kingma2013auto}. However, such a mechanism brings difficulties in modeling complex correlations at potentially high dimensions. Li~\etal \cite{li2024autoregressive} proposed Masked Auto-Regressive Models (\emph{MAR}), which brings flexible random generation orders by introducing masked auto-regressive ordering. MAR first transforms an image latent representation into n tokens $(x_1,...,x_n)$. Then it randomly masks some tokens in the sequence, and uses the unmasked tokens as conditions to train a lightweight diffusion head to model the denoising process of unmasked tokens. 

In inference, MAR uses a bi-level generation pattern: the outer auto-regressive loop controls the gradual unmasking process, which chooses to unmask some more masked tokens, while the inner diffusion chain uses the lightweight diffusion head conditioned on unmasked tokens to generate more tokens from masked ones. 
In the outer auto-regressive loop unmasking the latent tokens gradually, the \emph{masked auto-regressive} formulation randomly partitions token indices into subsets $S_1,\ldots,S_K$, predicting them group by group in K iterations:
\begin{align}
\label{eq:mar-infer}
p(x_1,...,x_n) = p(S_1,...,S_K) &= \prod_{k=1}^{K} p\big(S_k \mid S_{<k}\big) \nonumber \\
S_{<k} &= \bigcup_{j<k} S_j,
\end{align}
where $S_k = \left\{x_i, x_{i+1}..., x_j \right\}$ is a set of tokens to be predicted at the k-th step. 

The advantage of MAR diffusion is to benefit from the strengths of both sequence modeling and the diffusion models. However, such a bi-level generation pattern brings an unavoidable expensive computation cost, making MAR challenging in real-time generation applications. 



\subsection{One-step Diffusion and Preference Alignment}
\label{subsec:sim}
A major challenge of diffusion models is their slow multi-step sampling process, which requires hundreds to thousands of network evaluations. To address this, various distillation techniques have been proposed to accelerate sampling while preserving high-quality generation \cite{salimans2022progressive,song2023consistency,luo2024diff,yin2023one,zhou2024score,wang2025uni,kim2023consistency,songimproved,zhou2024long,heek2024multistep,xie2024distillation,salimans2024multistep,geng2024consistency,meng2022distillation,sauer2023adversarial,xu2024ufogen,zheng2024trajectory,li2024reward,berthelot2023tract,luo2023comprehensive}


\emph{Score Implicit Matching (SIM)} \cite{luo2024one} is a recent and powerful approach that enables one-step distillation of a pre-trained diffusion model without requiring access to real data during distillation. Unlike traditional knowledge distillation, which requires a dataset of teacher outputs, SIM operates by directly matching the score functions of the teacher and the student generator. Let the teacher's score function at noise level $t$ be $s_p(\boldsymbol{x},t)$, and the generator-induced score be $s_q(\boldsymbol{x},t)$. SIM minimizes a time-integrated score divergence between the two:
\begin{equation}
\mathcal{D}_d=\int_0^1 \lambda(t)\, E_{\boldsymbol{x}\sim \pi_t}\big[d\big(s_p(\boldsymbol{x},t),\, s_q(\boldsymbol{x},t)\big)\big]\ {\rm d}t,
\end{equation}
where $d(\cdot,\cdot)$ is a distance function (e.g., squared distance) and $\pi_t$ is a reference sampling distribution. A key insight of SIM is that while the generator's score $s_q$ is intractable, the gradient of this objective with respect to the generator's parameters $\theta$ can be efficiently computed. This is achieved by training an auxiliary network $\boldsymbol{S}_\phi(\boldsymbol{x_t},t)$ to approximate $s_q$, and then minimizing a tractable surrogate objective derived from the score-divergence gradient theorem ~\cite{luo2024one}.

The final training objective for the generator is:
\begin{align}
&\mathcal{L}_{SIM}(\theta) = \int_{t=0}^{T} w(t) 
\mathbb{E}_{\substack{z \sim p_z,\, x_0 = g_\theta(z) \\ x_t \mid x_0 \sim q_t(x_t \mid x_0)}} \left[l(\theta)\right] dt \\
&l(\theta)=\left\{ -\mathbf{d}'(y_t) \right\}^T \nonumber 
\left\{ s_{p_{sg[\theta]},t}(x_t) - \nabla_{x_t} \log q_t(x_t \mid x_0) \right\},
\end{align}
where $y_t:= s_{p_{sg[\theta]},t}(x_t) - s_{q_t}(x_t)$,  $g_\theta$ is the one-step generator and $\boldsymbol{z} \sim \mathcal{N}(0, I)$ is pure gaussian noise. The training alternates between updating the generator parameters $\theta$ using this objective and updating the auxiliary network $\boldsymbol{S}_\phi(\boldsymbol{x_t},t)$ with a standard denoising score matching loss. This dual-optimization framework allows SIM to effectively distill the knowledge of a multi-step teacher into a single-step generator in a data-free manner.

With the improved techniques of one-step diffusion, several works have explored practices of human preference alignment of diffusion models\cite{dai2023emu,podell2023sdxl,prabhudesai2023aligning,clark2023directly,lee2023aligning,fan2024reinforcement,black2023training,wallace2024diffusion,yang2024using,hong2024margin}. Particularly, some recent works \cite{luo2025reward,luo2024diffpp,luo2024david,luo2024diffstar} have advanced the human preference alignment for one-step diffusion models, with rigorous results. However, considering the introduction of random masked auto-regressive ordering, how to properly conduct preference alignment for one-step masked auto-regressive models remains unexplored.

\section{Method}
\label{sec:method}
\begin{figure*}[t]
  \centering
   \includegraphics[width=1.0 \linewidth]{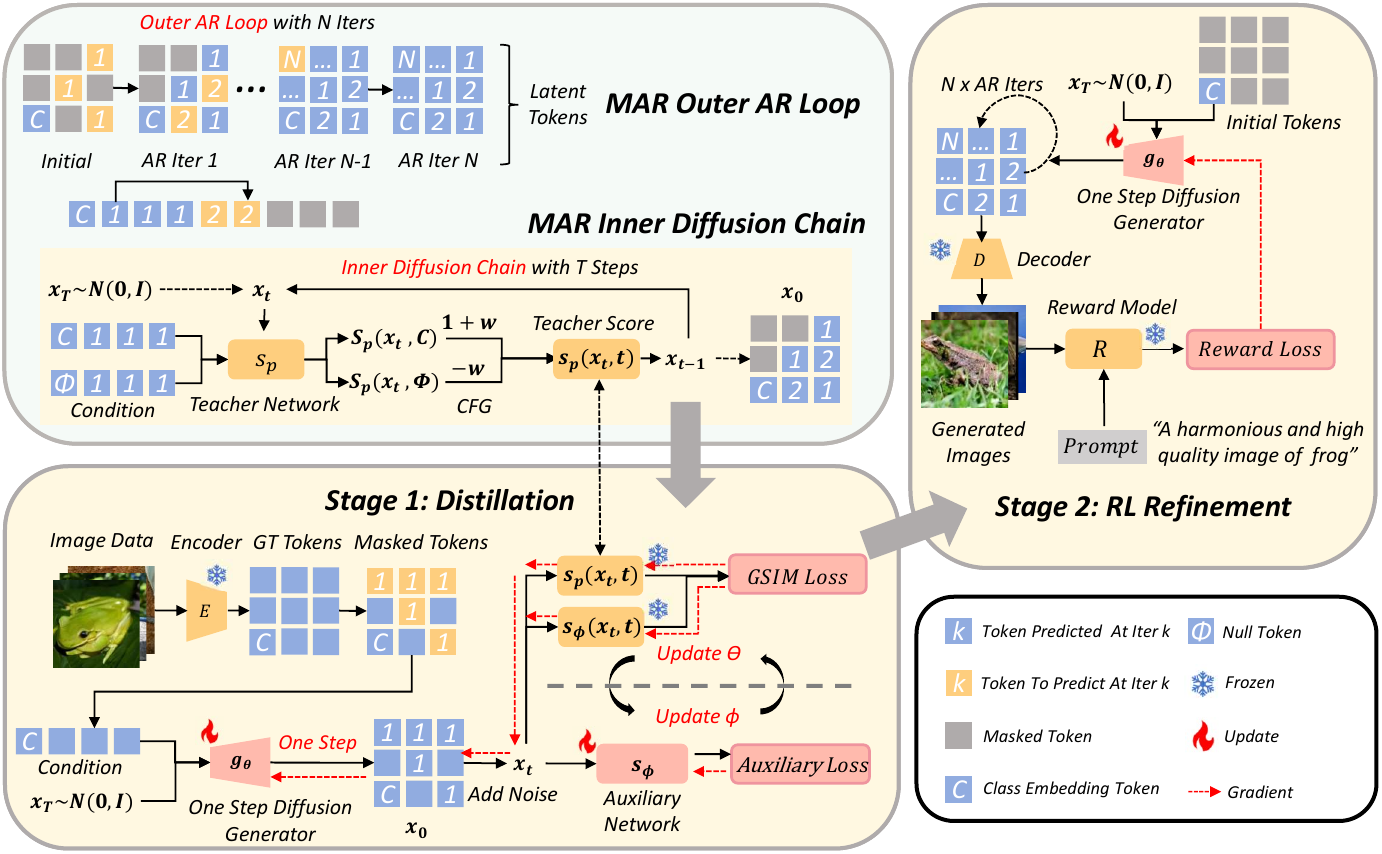} 
    \caption{Illustration of our overall framework. 
    \textbf{(Top-left)} The MAR inference process consists of an \textit{outer auto-regressive (AR) loop} and an \textit{inner diffusion chain}. 
    Starting from the class embedding token $c$, MAR performs multiple AR iterations, where each iteration predicts a new subset of latent tokens through a short diffusion process. 
    \textbf{(Bottom-left)} The student one-step generator $g_\theta$ and the auxiliary network are optimized alternately. In this stage, a portion of tokens is masked, and $g_\theta$ performs a single AR iteration to predict all masked tokens guided by the teacher MAR model's CFG-based predictions. 
    \textbf{(Right)} The RL refinement stage further improves perceptual fidelity. 
    Here, the distilled generator $g_\theta$ generates images through multi-step AR inference, and a reward model evaluates the outputs based on textual prompts. The reward loss then fine-tunes $g_\theta$ to better align with human perceptual preferences.}
    \label{fig:pipeline}
\end{figure*}

Our goal is to transform a slow, multi-step auto-regressive generation process, which we refer to as the \textbf{teacher}, into a fast, single-step \textbf{student} generator to enable the refinement of RL. We achieve this through a unified framework that first distills the conditional generation process in each outer AR loop and then fine-tunes the distilled model to align with human preferences. Our method consists of two main stages: (1) a novel distillation objective, Guided Score Implicit Matching (GSIM), to distill the multi-step pretrained diffusion denoising network into a single-step generator network $g_\theta$ by predicting the masked tokens from unmasked tokens following MAR's training process, and (2) 
a reinforcement learning stage that refines $g_\theta$ in the outer auto-regressive MAR loop using rewards from designed prompts and generated images. We describe our approach in the context of a Masked Auto-regressive (MAR) model, but the framework is generalizable to any AR+Diffusion architecture with any reward. Figure~\ref{fig:pipeline} shows the overall pipeline.

\subsection{Guided Score Implicit Matching (GSIM)}

To overcome the inference latency, our primary contribution is to distill the entire multi-step guided sampling process into a single-step generator, $g_\theta(z, c)$. Here, $z \sim p_z$ is a random noise vector sampled from prior distribution $p_z$, which is usually designated as the  standard gaussian distribution $\mathcal{N}(0, I)$. $c$ is the conditional auto-regressive context derived from unmasked tokens. The student generator $g_\theta$ is trained to directly map noise and context to a final, denoised tokens to predict.

While distilling the teacher’s multi-step guided sampler into a single-step student generator, we enforce distributional alignment between the student’s one-step process and the teacher’s multi-step process. We quantify the discrepancy with the KL divergence, leading to the objective:
\begin{equation}
\label{equ:kl-obj}
\begin{split}
    \theta^*:
    &=\arg \min_{\theta} \mathbb{E}_{c \sim \mathcal{C}} \left[ D_{\text{KL}}(q_\theta(\boldsymbol{x|c})\vert\vert p(\boldsymbol{x|c})) \right] ,
\end{split}
\end{equation}
where \(p(\mathbf{x}\mid c)\) is the teacher distribution, \(q_{\theta}(\mathbf{x}\mid c)\) is the student distribution, and \(c\in\mathcal{C}\) denotes the autoregressive context (e.g., class embeddings, unmasked tokens, and mask-position embeddings). However, directly optimizing the objective in Eq.~\ref{equ:kl-obj} is impossible, because the KL-loss of the two distribution is intractable.

Therefore, we introduce \textbf{Guided Score Implicit Matching (GSIM)}. Inspired by the research of Uni-instruct~\cite{wang2025uni}, the KL divergence is the integral of the Fisher divergence along a diffusion process under mild regularity conditions. Consequently, Eq.~\ref{equ:kl-obj} can be expressed in continuous time as the integral over \(t\) of the score discrepancy \(D_{\text{score}}\) between the student’s time-dependent score \(\nabla_{\mathbf{x}_t}\log q_{\theta,t}(\mathbf{x}_t \mid c)\) and the pretrained diffusion model’s score \(s_{p_t}(\mathbf{x}_t, c)\):
\begin{equation}
\begin{split}
& \mathcal{D}_{KL}(q_\theta(\boldsymbol{x|c})||p(\boldsymbol{x|c})) 
=\frac{1}{2}\int_0^T g^2(t)\mathbb{E}_{\boldsymbol{x}_t \sim q_{\theta,t}} \left[D_{\text{score}}(\boldsymbol{x}_t, \theta) \right] \text{d}t, \\
& D_{\text{score}} = ||\underbrace {\nabla_{\boldsymbol{x}_t} \log q_{\theta,t}(\boldsymbol{x}_t \mid c)}_{\text{student score}} - \underbrace{s_{p_t}(\boldsymbol{x}_t, c)}_{\text{teacher score}} ||^2_2,
\label{eq:uni_score_divergence}
\end{split}
\end{equation}
where $g(t)$ is the diffusion coefficient scaling Brownian motion in Eq.~\ref{eq:sde}. The single-step student score $\nabla_{\boldsymbol{x}_t} \log q_{\theta,t}(\boldsymbol{x}_t \mid c)$ is approximated by an auxiliary network, $\boldsymbol{S}_\phi(\boldsymbol{x}_t, t, c)$, which is trained on the current set of reconstructions $x\sim q_\theta(\boldsymbol{x})$ using a standard denoising score matching objective:
\begin{equation}
\label{eq:online_loss}
\begin{split}
&\mathcal{L}_{auxiliary}(\phi) = \int_0^T \lambda(t)\mathbb{E}_{\substack{ 
        \boldsymbol{x}_0 \sim q_{\theta,0} \\ 
        \boldsymbol{x}_t | \boldsymbol{x}_0 \sim p_t(\boldsymbol{x}_t | \boldsymbol{x}_0)
        }} \left[ l(\phi) \right] dt,  \\ 
& l(\phi) =\left\| \boldsymbol{S}_\phi(\mathbf{x}_t, t, c) - \nabla_{\mathbf{x}_t}\log p_{t}(\mathbf{x}_t|\mathbf{x}_0, c) \right\|_2^2.
\end{split}
\end{equation}
Since MAR employs \textbf{classifier-free guidance (CFG)}~\cite{ho2022classifier} during inference, the teacher score $s_{p_t}(\boldsymbol{x}_t, c)$ in Eq.~\ref{eq:uni_score_divergence} must correspond to the CFG-guided diffusion process. Accordingly, we define the guided teacher score as:
\begin{equation}
\label{eq:modified_cfg}
s_{p_t}(\boldsymbol{x}_t, c) = (1+w) \cdot \boldsymbol{S}_p(\boldsymbol{x}_t, t, c) - w \cdot \boldsymbol{S}_p(\boldsymbol{x}_t, t, \emptyset),
\end{equation}
where $w$ is the guidance scale, $\emptyset$ denotes the null condition, and $\boldsymbol{S}_p$ is the neural network approximation of the teacher's score. This ensures that the student model learns from the CFG-guided distribution that the teacher actually samples during MAR inference. With all terms in Eq.~\ref{eq:uni_score_divergence} defined, we now optimize the student parameters $\theta$ using the following gradient equivalence theorem.
\begin{customtheorem}[Gradient Equivalent Theorem]
    The symbol $sg[\cdot]$ means stop gradient of some parameters. If distribution \( q_\theta(\boldsymbol{x}) \) satisfies some mild regularity conditions, we have the theorem for any score function \( s_{p_t}(\cdot) \):
    \begin{align*}
        & \mathbb{E}_{\boldsymbol{x}_t \sim q_{\theta,t}}  \frac{\partial}{\partial\theta} d \left( s_{q_{\theta,t}}(\boldsymbol{x}_t, c) - s_{p_t}(\boldsymbol{x}_t, c) \right) \\ 
        &=\mathbb{E}_{\boldsymbol{x}_t \sim q_{\theta,t}} \Biggl[ d' \left( s_{q_{\theta,t}}(\boldsymbol{x}_t, c) - s_{p_t}(\boldsymbol{x}_t, c) \right) \frac{\partial}{\partial\theta} s_{q_{\theta,t}}(\boldsymbol{x}_t, c) \\ 
        &+ \frac{\partial}{\partial \boldsymbol{x}_t }d(s_{q_{\theta,t}}(\boldsymbol{x}_t, c) - s_{p_t}(\boldsymbol{x}_t, c)) \frac{\partial \boldsymbol{x}_t }{\partial \theta } \Biggr] \\ 
        &= \frac{\partial}{\partial \theta} \mathbb{E}_{\substack{ 
        \boldsymbol{x}_0 \sim q_{\theta,0} \\
        \boldsymbol{x}_t | \boldsymbol{x}_0 \sim p_t(\boldsymbol{x}_t | \boldsymbol{x}_0)
        }} 
        \Biggl[ -\left\{ d' \left( s_{q_{\text{sg}[\theta],t}}(\boldsymbol{x}_t, c) - s_{p_t}(\boldsymbol{x}_t, c) \right) \right\}^T \\ 
         &\left\{ s_{q_{\text{sg}[\theta],t}}(\boldsymbol{x}_t, c) - \nabla_{\boldsymbol{x}_t} \log p_t(\boldsymbol{x}_t | \boldsymbol{x}_0, c) \right\} \\ 
         & + d(s_{q_{\text{sg}[\theta],t}}(\boldsymbol{x}_t, c) - s_{p_t}(\boldsymbol{x}_t, c))\Biggr]
    \end{align*}
    Please refer to supplementary materials for the proof.
\end{customtheorem}

According to this theorem, The optimization problem in Eq.~\ref{equ:kl-obj} and \ref{eq:uni_score_divergence} can be formulated as
\begin{equation}
\label{eq:gsim}
\begin{split}
    &\theta^*:
    =\arg \min_{\theta} \mathbb{E}_{\substack{ 
        c \sim \mathcal{C},\space z \sim p_z, \space  t\sim\mathcal{U}[0,1] \\
        \boldsymbol{x}_0 = g_{\theta}(z, c),\
        \boldsymbol{x}_t | \boldsymbol{x}_0 \sim p_t(\boldsymbol{x}_t | \boldsymbol{x}_0)
        }} \left[ L_{GSIM}(\theta) \right], \\ 
    &L_{GSIM} = L_1 + L_2, \\ 
    &L_1 = - \left\{ d' \left( y_t \right) \right\}^T 
         \left\{ s_{q_{\text{sg}[\theta],t}}(\boldsymbol{x}_t, c) - \nabla_{\boldsymbol{x}_t} \log p_t(\boldsymbol{x}_t | \boldsymbol{x}_0, c) \right\}, \\ 
    &L_2 =d(y_t), \quad y_t = s_{q_{\text{sg}[\theta],t}}(\boldsymbol{x}_t, c) -  s_{p_t}(\boldsymbol{x}_t, c), \\ 
\end{split}
\end{equation}
where $s_{q_{\text{sg}[\theta],t}}(\boldsymbol{x}_t, c)$ can be approximated by the auxiliary network $\boldsymbol{S}_\phi(\boldsymbol{x}_t, t, c)$. Although theory suggests using the squared \(\ell_2\) distance \( \|\cdot\|_2^2 \) for \( d(\cdot,\cdot) \) in Eq.~\ref{eq:uni_score_divergence} is necessary to recover the KL objective exactly, in practice this choice can cause training instability. For improved training robustness and faster convergence, we adopt the \emph{Pseudo-Huber distance} as the loss function in our implementation, following~\cite{luo2024one}. Concretely, in Eq.~\ref{eq:gsim} we set
\begin{equation}
\begin{split}
    d(y_t) &= \sqrt{||y_t||^2_2+ r^2} - r, \\
    d'(y_t) &= y_t / \sqrt{||y_t||^2_2+ r^2}
\end{split}
\end{equation}
where r is a small constant to avoid division by zero (we use $r=1e-5$).

\subsection{Reinforcement Learning for Post-Hoc Refinement}
\label{RL}
Although GSIM efficiently matches the teacher’s output distribution, that distribution can still diverge from human perceptual preferences—primarily due to cumulative errors across the autoregressive loops. To further improve perceptual fidelity, we introduce a reinforcement learning (RL) fine-tuning stage.

However, the RL process cannot be seamlessly integrated with the GSIM distillation phase. The key reason lies in the fundamental difference between MAR’s training dynamics and its inference process. During both MAR training and GSIM distillation, the model operates with a fraction of tokens kept unmasked and performs only a \textbf{single} auto-regressive (AR) iteration, where the diffusion model predicts all masked tokens conditioned on the visible context $c$. This single-step generation, while sufficient for learning semantic consistency from the teacher, produces low-fidelity images that lack fine perceptual details.

In contrast, MAR inference begins solely from the class embedding $c_{emb}$ and relies on \textbf{multiple} AR iterations to progressively refine image quality. If we were to couple RL with GSIM and directly optimize on single-step samples, the reward model would evaluate these low-quality intermediate outputs rather than the final multi-iteration results. This would lead to misleading gradients and unstable policy optimization, since the reward signal would not correspond to the perceptual quality achieved during inference.

Therefore, RL fine-tuning must be formulated as a separate, \textit{post-hoc} refinement phase applied to the distilled generator $g_\theta$. In this stage, $g_\theta$ performs multi-step AR generation consistent with MAR inference, ensuring that the reward feedback reflects the true perceptual quality of the final outputs.

We treat the single-step generator $g_\theta$ as a policy. For a given class $c_{emb}$, the policy generates an output with $K$ iterations as illustrated in Eq.~\ref{eq:mar-infer}, $x_g = G_\theta(z, c_{emb}, K)$, which we consider an "action." We use $G_\theta$ to differ the multi-iteration generation from single-iteration $g_\theta$. To guide this policy, we employ a pre-trained reward model that provides a scalar score reflecting the alignment between the generated image and the textual context. For specific $c_{emb}$, we can get the class name in text, such as 'frog', so we can design the prompt $prompt_c$ about corresponding class like '\emph{a high-quality and harmonious picture of \{class name\}}'. Let this reward function be denoted by $R(x_g, prompt_c)$. The RL objective is to maximize the expected reward of the generated outputs, which is achieved by minimizing the following loss function:
\begin{equation}
\label{eq:rl_loss}
\mathcal{L}_{\text{RL}} = - \mathbb{E}_{c_{emb}, z} \left[ R(G_\theta(z, c_{emb}, K), prompt_c) \right]
\end{equation}
By back propagating the gradient from this reward signal, we fine-tune $\theta$ to produce outputs that are not only consistent with the auto-regressive prior but also explicitly optimized for higher quality and better text-image alignment. For our specific implementation, we utilize \textbf{PickScore}~\cite{kirstain2023pick} as the reward model $R(\cdot)$, as it is a powerful function known to correlate well with human aesthetic judgments.

\section{Experiments}
\label{sec:experiments}

In this section, we evaluate the proposed distillation and reinforcement learning (RL) scheme for our MARVAL (Masked Auto-regressive Variational Acceleration). We conduct four sets of experiments to investigate (1) the influence of classifier-free guidance (CFG) scales on the distilled model, (2) the acceleration achieved over MAR under varying numbers of autoregressive iterations and diffusion steps, (3) the overall effectiveness and significance of RL refinement across models of different scales, and (4) the generalization capability of our framework to text-to-image generation tasks.

\subsection{Experimental Setup}

Following MAR~\cite{li2024autoregressive}, our diffusion process adopts a cosine noise schedule with 1000 training steps. The denoising network predicts the noise vector $\varepsilon$ through a lightweight MLP.
We evaluate three model scales identical to MAR: \textbf{Base} (6 blocks, 1024 width, 208M params), \textbf{Large} (8 blocks, 1280 width, 479M params), and \textbf{Huge} (12 blocks, 1536 width, 943M params).
In the distillation stage, we train each model on 8 NVIDIA A800 GPUs for 30 epochs, which takes approximately 3 days to complete. In the RL refinement stage, we further refine the distilled model on the same hardware configuration for 5 epochs, requiring about 2 days of training.


For robust evaluation, class-conditional metrics are evaluated on 50K samples uniformly distributed across 1,000 ImageNet classes, while text-to-image metrics use 50K prompts randomly sampled from the JourneyDB~\cite{sun2023journeydb} test set.

\subsection{Classifier-Free Guidance Scale in Distillation}
According to Eq.~\ref{eq:modified_cfg}, we set the Classifier-Free Guidance (CFG) scale $w$ manually during the distillation stage. To study its effect, we vary $w$ for the base MAR model (MAR-B) and evaluate the distilled results using FID and IS.

\begin{table}[h]
\centering
\begin{tabular}{lccc}
\toprule
Model & CFG scale ($w$) & FID$\downarrow$ & IS$\uparrow$ \\
\midrule
MARVAL-B & $w=1$ (w/o CFG) & 3.75 & 182.1 \\
 &$w=1.2$ & 3.06 & 220.2 \\
 &$w=1.5$ & 4.24 & 261.7 \\
 &$w=2$ & 7.98 & 294.4 \\
 &$w=3$ & 11.98 & \textbf{317.1} \\
 &$w=4$ & 13.84 & 316.8 \\
\midrule
MAR-B &$w=2.9$ & \textbf{2.60} & 222.7 \\
\bottomrule
\end{tabular}
\caption{FID and IS of distilled one-step models under different CFG scales, compared with the optimal setting of MAR-B.}
\label{tab:cfg}
\end{table}


As shown in Table~\ref{tab:cfg}, increasing $w$ improves IS, indicating enhanced image fidelity, but overly large $w$ degrades FID, which suggests sacrificing the diversity and the data alignment for better fidelity. Since ImageNet contains low-quality samples, we aim to balance realism and fidelity rather than minimizing FID alone.

We therefore select $w=1.2$ as the optimal setting. This choice yields a favorable trade-off between visual sharpness and distribution consistency.


\subsection{Effect of Distillation}
\begin{figure}[!h]
\centering
\includegraphics[width=1\linewidth]{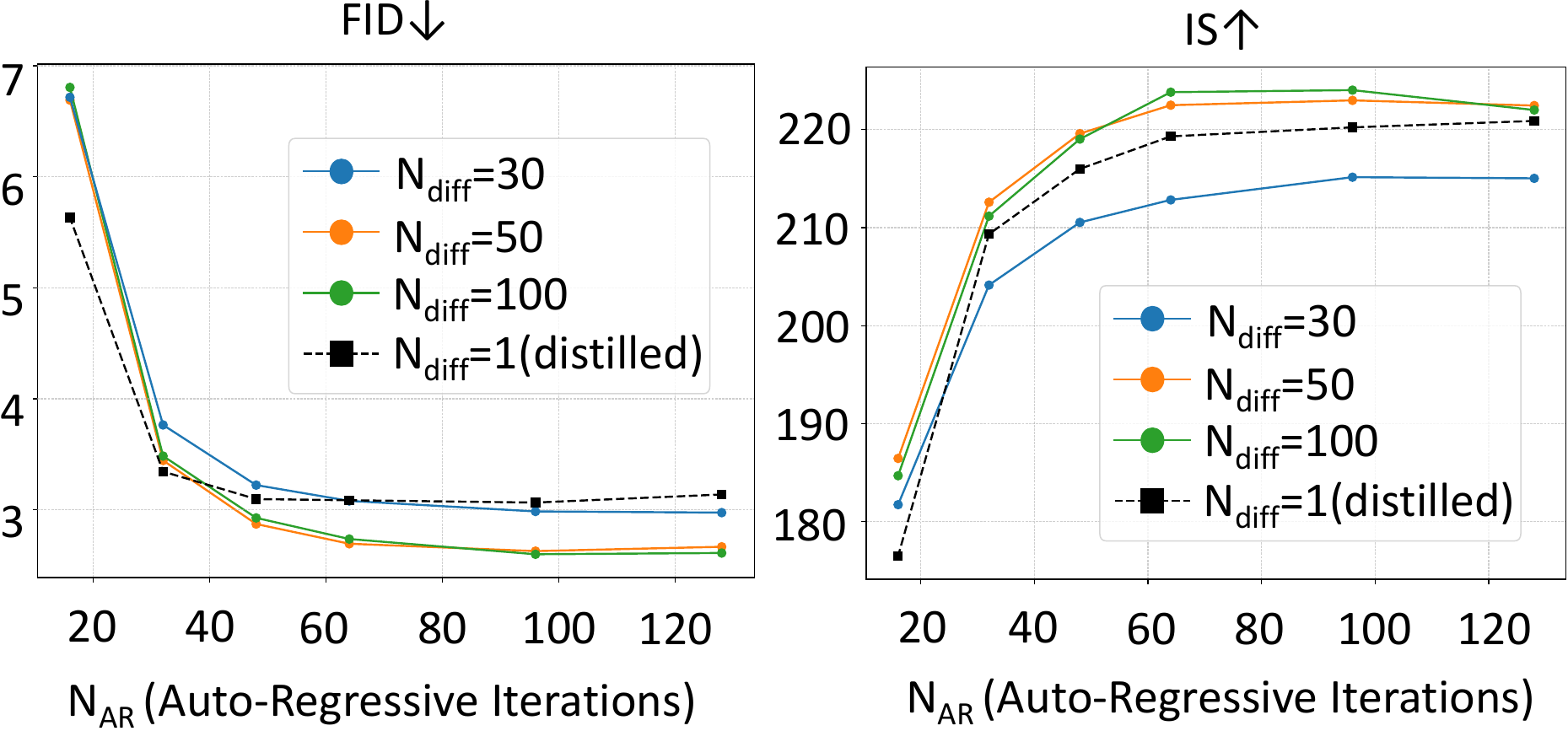}
\caption{Effect of AR iterations and diffusion steps on FID/IS for the MAR base model compared to MARVAL base model.}
\label{fig:base_fid_is}
\end{figure}

We further investigate the effect of distillation on the \textbf{Base model} by varying the number of auto-regressive iterations ($N_{\text{AR}}$) and diffusion steps ($N_{\text{diff}}$). As shown in Fig.~\ref{fig:base_fid_is}, increasing $N_{\text{diff}}$ gradually improves both FID and IS, while the performance saturates around $N_{\text{diff}}=100$, consistent with observations in the MAR paper.

After distillation, the one-step model achieves performance comparable to MAR-B under its optimal configuration ($N_{\text{diff}}=100$) and even surpasses MAR-B with $N_{\text{diff}}=30$. This indicates that our distillation effectively preserves the teacher model’s generative capability while dramatically simplifying the sampling process.

\FloatBarrier
\begin{figure*}[h]
    \centering
    \setlength{\tabcolsep}{1pt}
    \setlength{\fboxrule}{1pt}
    \resizebox{0.98\textwidth}{!}{
    \begin{tabular}{c}
    \begin{tabular}{ccccccc}
        & golden retriever & cougar & hare & dowitcher & Pekinese dog & diamondback \\
        \begin{turn}{90} \,\,\,\,\,\,\,\,\,\,\,\,\,\,\,\,\,\,\,\small{{MAR-B}} \end{turn} &
        \includegraphics[width=0.2\textwidth]{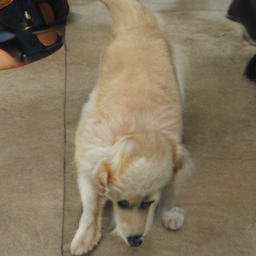} &
        \includegraphics[width=0.2\textwidth]{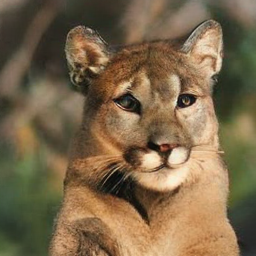} &
        \includegraphics[width=0.2\textwidth]{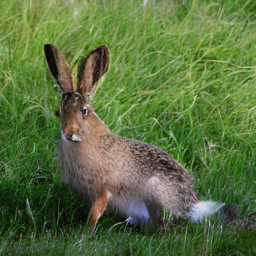} &
        \includegraphics[width=0.2\textwidth]{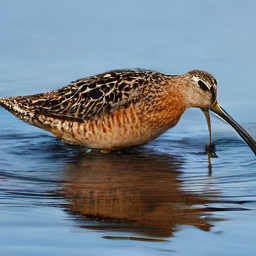} &
        \includegraphics[width=0.2\textwidth]{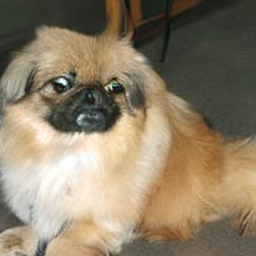} &
        \includegraphics[width=0.2\textwidth]{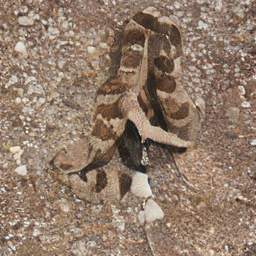} \\
        \begin{turn}{90} \,\,\,\,\,\,\,\,\,\,\,\,\,\,\,\small{{MARVAL-B}} \end{turn} &
        \includegraphics[width=0.2\textwidth]{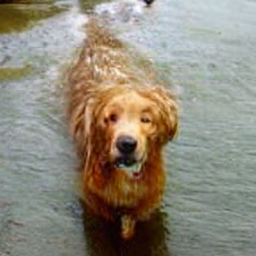} &
        \includegraphics[width=0.2\textwidth]{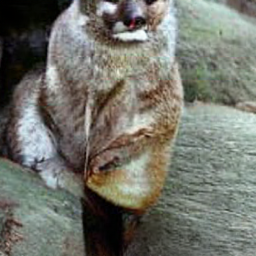} &
        \includegraphics[width=0.2\textwidth]{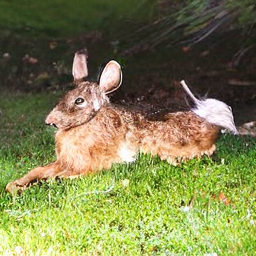} &
        \includegraphics[width=0.2\textwidth]{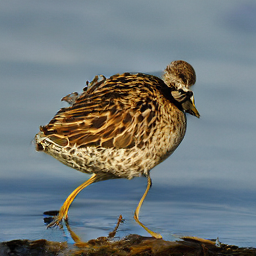} &
        \includegraphics[width=0.2\textwidth]{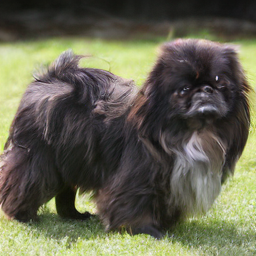} &
        \includegraphics[width=0.2\textwidth]{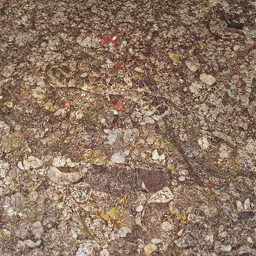} \\
        \begin{turn}{90} \,\,\,\,\,\,\,\,\,\,\,\small{{MARVAL-RL-B}} \end{turn} &
        \includegraphics[width=0.2\textwidth]{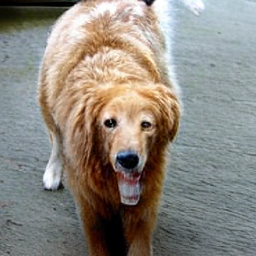} &
        \includegraphics[width=0.2\textwidth]{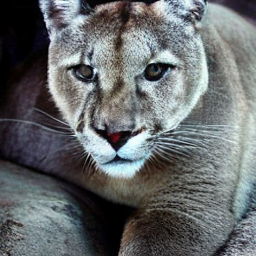} &
        \includegraphics[width=0.2\textwidth]{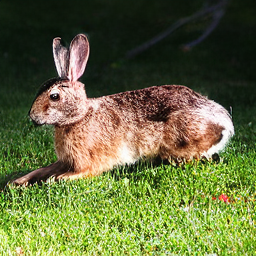} &
        \includegraphics[width=0.2\textwidth]{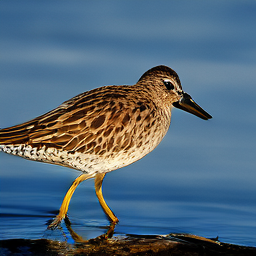} &
        \includegraphics[width=0.2\textwidth]{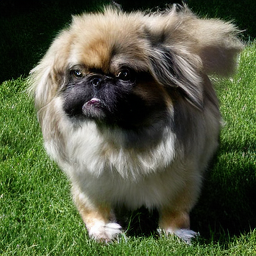} &
        \includegraphics[width=0.2\textwidth]{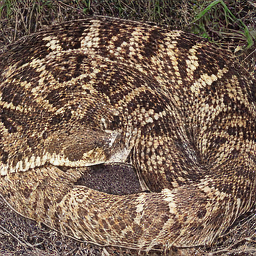} \\
    \end{tabular}
    \end{tabular}}
    \caption{\textbf{Qualitative comparison among MAR-B, MARVAL-B (without RL), and MARVAL-RL-B.}The \textbf{first row (MAR-B)} shows MAR-B generation results, which capture coarse object semantics but often suffer from limited visual realism, noticeable artifacts, and poor texture fidelity. 
    The \textbf{second row (MARVAL-B)} shows that images generated by the distilled one-step model tend to be smoother and less detailed—animal faces appear blurred or distorted, and fine-grained textures or contextual cues are frequently lost. 
    The \textbf{third row (MARVAL-RL-B)} demonstrates the effect of reinforcement learning refinement: details such as fur texture, feather edges, and body contours become significantly sharper and more coherent. The enhanced structure and perceptual realism highlight RL as an essential stage for aligning the model’s outputs with human visual preferences.
    }
    \label{fig:compare}
    \vspace{-0.1em}
\end{figure*}

We also evaluate the \textbf{Large} and \textbf{Huge} variants, summarized in Table~\ref{tab:fidis_distill}. The MARVAL models maintain better FID than MAR while achieving significantly higher Inception Scores, confirming that distillation improves perceptual quality without compromising data alignment. The FID and IS in Table~\ref{tab:fidis_distill} of MAR and MARVAL are achieved with the optimal settings on NVIDIA
A100 GPUs, and the inference time is tested when batch size equals 1 on a single A100 GPU. For the \textbf{Base} model, the generation takes only \textbf{0.61 seconds} after distillation, compared to \textbf{20.10 seconds} of MAR-B. The diffusion sampling process itself is accelerated by nearly \textbf{100×}, and considering the auto-regressive component, the overall real time generation speed improves by about \textbf{32.95×}. Overall, our MARVAL models perform the best in generation quality and speed.

\begin{table}[h]
\centering
\begin{tabular}{lccc}
\toprule
Model & FID$\downarrow$ & IS$\uparrow$ & inference time \\
\midrule
\multicolumn{4}{l}{\textit{pixel-based}} \\
ADM~\cite{dhariwal2021diffusion} & 4.59 & 186.7 & 115.67 s\\
\midrule
\multicolumn{4}{l}{\textit{continuous-valued tokens}} \\
LDM-4~\cite{rombach2022high} & 3.60 & 247.7 & 37.25s\\
U-ViT-H/2~\cite{bao2022all} & 2.29 & 263.9 & 3.01 s\\
DiT-XL/2~\cite{peebles2023scalable} & 2.27 & \textbf{278.2} & 6.62 s\\
\midrule
\multicolumn{4}{l}{\textit{masked auto-regressive}} \\
MAR-B ($w=2.9$)& 2.60 & 222.7 & 20.10 s \\
MAR-L ($w=3.0$)& 2.23 & 240.5 & 24.04 s \\
MAR-H ($w=3.2$)& 2.06 & 247.7 & 30.98 s \\
\midrule
\multicolumn{4}{l}{\textit{distilled only}} \\
MARVAL-B & 3.06 & 220.2 & \textbf{0.61} s\\
MARVAL-L & 2.51 & 247.1 & 0.97 s\\
MARVAL-H & \textbf{2.00} & 256.3 & 1.67 s\\
\bottomrule
\end{tabular}
\caption{System-level comparison on ImageNet 256×256 conditional generation. 
Our MARVAL model achieves the best performance in FID while requiring the least inference time. All MARVAL models are evaluated with a CFG scale of 1.2, whereas MAR baselines use the CFG settings provided in their official evaluation scripts.}
\label{tab:fidis_distill}
\end{table}

\subsection{The Significance of Reinforcement Learning}  

To assess the impact of reinforcement learning (RL) fine-tuning, we perform an ablation study comparing the distilled models before and after RL optimization. RL fine-tuning is applied to better align image generation with perceptual and semantic objectives. RL markedly improves visual quality and perceptual realism, as reflected in the significant improvent of reward-based metrics in Table~\ref{tab:reward_rl}.

\begin{table}[h]
\centering
\begin{tabular}{lcccc}
\toprule
Model & \multicolumn{2}{c}{Distill w/o RL} & \multicolumn{2}{c}{Distill w RL} \\
\cmidrule(lr){2-3} \cmidrule(lr){4-5}
& CLIP$\uparrow$ & ImgR$\uparrow$ & CLIP$\uparrow$ & ImgR$\uparrow$ \\
\midrule
Base  & 29.40 & -0.142 & 29.47 & -0.110 \\
Large & 29.43 & -0.113 & 29.84 & -0.076 \\
Huge  & 29.33 & -0.107 & 29.78 & -0.064  \\
\bottomrule
\end{tabular}
\caption{Reward-based evaluation (CLIP and ImageReward) before and after RL fine-tuning. Higher scores indicate better semantic fidelity and perceptual alignment with human preference.}
\label{tab:reward_rl}
\end{table}

As illustrated in Fig.~\ref{fig:compare}, RL fine-tuning brings clear improvements across diverse object categories. 
The \textbf{MAR-B} baseline captures coarse semantics but often produces unrealistic textures and structural distortions—examples such as the \textit{hare} and \textit{diamondback} exhibit oversimplified shapes. 
Without RL, examples such as the \textit{hare}, \textit{cougar}, and \textit{dowitcher} appear blurry and distorted, while the \textit{diamondback} loses clear boundaries. 
After RL, main subjects become more salient, contours and textures are better preserved, and the overall composition appears more natural. 
These results confirm that RL guides the model toward sharper, semantically faithful, and perceptually coherent outputs, successfully enhancing the distilled model despite minor metric degradations.
Overall, these results confirm that RL fine-tuning effectively enhances image fidelity and perceptual quality, demonstrating the success of our approach.

\subsection{Generalization to Text-to-Image Generation}

To demonstrate our framework's versatility beyond class-conditional generation, we apply our MARVAL-RL framework to the Text-to-Image (T2I) DC-AR model~\cite{wu2025dc}.

T2I tasks require high-fidelity synthesis and strict prompt alignment. Evaluated on 50,000 samples (Table~\ref{tab:dcar_reward}), our 1-step+RL strategy compresses the original 20-step DDIM sampling into a single step. The compression does not degrade performance; instead, RL fine-tuning significantly boosts the ImageReward score from 0.5876 to 0.6903, highlighting RL's value in complex alignment tasks.

\begin{figure}[h]
\centering
\includegraphics[width=0.8\linewidth]{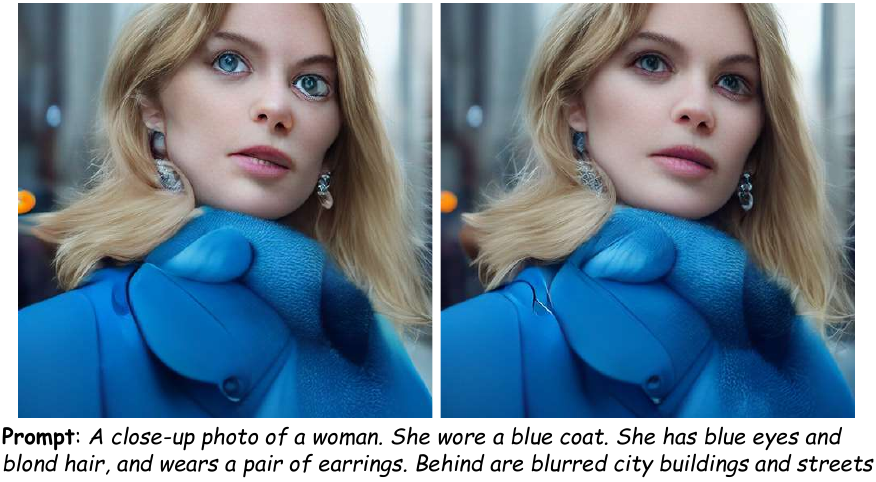} 
\caption{T2I qualitative comparison. \textbf{Left:} Original 20-step DC-AR. \textbf{Right:} Our 1-step+RL DC-AR. The RL-finetuned one-step model exhibits enhanced perceptual quality and visual harmony.}
\label{fig:DC-AR}
\end{figure}

Qualitatively (Figure~\ref{fig:DC-AR}), the 1-step+RL DC-AR model maintains excellent visual and semantic fidelity. It accurately captures fine-grained details (e.g., facial structure, hair texture) and adheres strictly to the prompt, proving our framework successfully preserves generative quality even under extreme acceleration. Additional high-quality generation results are provided in the supplementary material. 

\begin{table}[h]
\centering
\begin{tabular}{lcc}
\toprule
Model (Text-to-Image) & Strategy & ImageReward $\uparrow$ \\
\midrule
DC-AR-512 & 20-step DDIM & 0.5876 \\
DC-AR (distilled+RL) & \textbf{1-step + RL} & \textbf{0.6903} \\
\bottomrule
\end{tabular}
\caption{ImageReward on 50K T2I samples. Our strategy compresses inference while significantly boosting alignment.}
\label{tab:dcar_reward}
\end{table}

\section{Conclusion}
In this work, we addressed the fundamental limitations of Masked Auto-Regressive (MAR) models—slow inference and the infeasibility of fine-tuning with reinforcement learning—by proposing a unified variational acceleration framework, GSIM. Built upon the novel score-based variational principle, our method distills the costly diffusion denoising chains in MAR into a single-step generator, dramatically improving inference efficiency while preserving the auto-regressive ordering.

This acceleration not only alleviates the computational bottleneck of MAR models but also enables effective post-training with reinforcement learning, which was previously impractical due to excessive computational costs. Leveraging human preference rewards, our MARVAL-RL model achieves both over 30× acceleration and better alignment with perceptual quality and semantic fidelity compared to conventional MAR diffusion decoding.




\section{Acknowledgement}

This work was supported by the Shanghai Municipal Science and Technology Major Project (2025SHZDZX026D03), the National Key Research and Development Program of China (2024YFC3406400), and the National Natural Science Foundation of China (32450631, 62371007). Additionally, we acknowledge the support from the Biomedical Computing Platform of National Biomedical Imaging Center, Peking University.
{ 
    \small
    \bibliographystyle{ieeenat_fullname}
    \bibliography{main}

@String(ICCV= {Int. Conf. Comput. Vis.})

@String(ICCV  = {ICCV})

@article{ho2020denoising,
  title={Denoising diffusion probabilistic models},
  author={Ho, Jonathan and Jain, Ajay and Abbeel, Pieter},
  journal={Advances in neural information processing systems},
  volume={33},
  pages={6840--6851},
  year={2020}
}

@article{song2020denoising,
  title={Denoising diffusion implicit models},
  author={Song, Jiaming and Meng, Chenlin and Ermon, Stefano},
  journal={arXiv preprint arXiv:2010.02502},
  year={2020}
}

@article{song2019generative,
  title={Generative modeling by estimating gradients of the data distribution},
  author={Song, Yang and Ermon, Stefano},
  journal={Advances in neural information processing systems},
  volume={32},
  year={2019}
}

@article{song2020score,
  title={Score-based generative modeling through stochastic differential equations},
  author={Song, Yang and Sohl-Dickstein, Jascha and Kingma, Diederik P and Kumar, Abhishek and Ermon, Stefano and Poole, Ben},
  journal={arXiv preprint arXiv:2011.13456},
  year={2020}
}

@article{li2024autoregressive,
  title={Autoregressive image generation without vector quantization},
  author={Li, Tianhong and Tian, Yonglong and Li, He and Deng, Mingyang and He, Kaiming},
  journal={Advances in Neural Information Processing Systems},
  volume={37},
  pages={56424--56445},
  year={2024}
}

@article{luo2024one,
  title={One-step diffusion distillation through score implicit matching},
  author={Luo, Weijian and Huang, Zemin and Geng, Zhengyang and Kolter, J Zico and Qi, Guo-jun},
  journal={Advances in Neural Information Processing Systems},
  volume={37},
  pages={115377--115408},
  year={2024}
}

@article{kingma2013auto,
  title={Auto-encoding variational bayes},
  author={Kingma, Diederik P and Welling, Max},
  journal={arXiv preprint arXiv:1312.6114},
  year={2013}
}

@article{salimans2022progressive,
  title={Progressive Distillation for Fast Sampling of Diffusion Models},
  author={Salimans, Tim and Ho, Jonathan},
  journal={arXiv preprint arXiv:2202.00512},
  year={2022}
}

@article{song2023consistency,
  title={Consistency models},
  author={Song, Yang and Dhariwal, Prafulla and Chen, Mark and Sutskever, Ilya},
  year={2023}
}

@inproceedings{zhou2024score,
  title={Score identity distillation: Exponentially fast distillation of pretrained diffusion models for one-step generation},
  author={Zhou, Mingyuan and Zheng, Huangjie and Wang, Zhendong and Yin, Mingzhang and Huang, Hai},
  booktitle={Forty-first International Conference on Machine Learning},
  year={2024}
}

@article{ho2022classifier,
  title={Classifier-free diffusion guidance},
  author={Ho, Jonathan and Salimans, Tim},
  journal={arXiv preprint arXiv:2207.12598},
  year={2022}
}

@article{salimans2016improved,
  title={Improved techniques for training gans},
  author={Salimans, Tim and Goodfellow, Ian and Zaremba, Wojciech and Cheung, Vicki and Radford, Alec and Chen, Xi},
  journal={Advances in neural information processing systems},
  volume={29},
  year={2016}
}

@article{kirstain2023pick,
  title={Pick-a-pic: An open dataset of user preferences for text-to-image generation},
  author={Kirstain, Yuval and Polyak, Adam and Singer, Uriel and Matiana, Shahbuland and Penna, Joe and Levy, Omer},
  journal={Advances in neural information processing systems},
  volume={36},
  pages={36652--36663},
  year={2023}
}

@inproceedings{rombach2022high,
  title={High-resolution image synthesis with latent diffusion models},
  author={Rombach, Robin and Blattmann, Andreas and Lorenz, Dominik and Esser, Patrick and Ommer, Bj{\"o}rn},
  booktitle={Proceedings of the IEEE/CVF conference on computer vision and pattern recognition},
  pages={10684--10695},
  year={2022}
}

@inproceedings{peebles2023scalable,
  title={Scalable diffusion models with transformers},
  author={Peebles, William and Xie, Saining},
  booktitle={Proceedings of the IEEE/CVF international conference on computer vision},
  pages={4195--4205},
  year={2023}
}

@inproceedings{bao2022all,
  title={All are worth words: a vit backbone for score-based diffusion models},
  author={Bao, Fan and Li, Chongxuan and Cao, Yue and Zhu, Jun},
  booktitle={NeurIPS 2022 Workshop on Score-Based Methods},
  year={2022}
}

@article{dhariwal2021diffusion,
  title={Diffusion models beat gans on image synthesis},
  author={Dhariwal, Prafulla and Nichol, Alexander},
  journal={Advances in neural information processing systems},
  volume={34},
  pages={8780--8794},
  year={2021}
}

@inproceedings{deng2009imagenet,
  title={Imagenet: A large-scale hierarchical image database},
  author={Deng, Jia and Dong, Wei and Socher, Richard and Li, Li-Jia and Li, Kai and Fei-Fei, Li},
  booktitle={2009 IEEE conference on computer vision and pattern recognition},
  pages={248--255},
  year={2009},
  organization={Ieee}
}

@article{hoogeboom2021autoregressive,
  title={Autoregressive diffusion models},
  author={Hoogeboom, Emiel and Gritsenko, Alexey A and Bastings, Jasmijn and Poole, Ben and Berg, Rianne van den and Salimans, Tim},
  journal={arXiv preprint arXiv:2110.02037},
  year={2021}
}

@inproceedings{van2016pixel,
  title={Pixel recurrent neural networks},
  author={Van Den Oord, A{\"a}ron and Kalchbrenner, Nal and Kavukcuoglu, Koray},
  booktitle={International conference on machine learning},
  pages={1747--1756},
  year={2016},
  organization={PMLR}
}

@article{christiano2017deep,
  title={Deep reinforcement learning from human preferences},
  author={Christiano, Paul F and Leike, Jan and Brown, Tom and Martic, Miljan and Legg, Shane and Amodei, Dario},
  journal={Advances in neural information processing systems},
  volume={30},
  year={2017}
}

@article{xue2023sa,
  title={Sa-solver: Stochastic adams solver for fast sampling of diffusion models},
  author={Xue, Shuchen and Yi, Mingyang and Luo, Weijian and Zhang, Shifeng and Sun, Jiacheng and Li, Zhenguo and Ma, Zhi-Ming},
  journal={Advances in Neural Information Processing Systems},
  volume={36},
  pages={77632--77674},
  year={2023}
}

@article{lu2022dpm,
  title={DPM-Solver: A Fast ODE Solver for Diffusion Probabilistic Model Sampling in Around 10 Steps},
  author={Lu, Cheng and Zhou, Yuhao and Bao, Fan and Chen, Jianfei and Li, Chongxuan and Zhu, Jun},
  journal={arXiv preprint arXiv:2206.00927},
  year={2022}
}

@article{karras2022elucidating,
  title={Elucidating the design space of diffusion-based generative models},
  author={Karras, Tero and Aittala, Miika and Aila, Timo and Laine, Samuli},
  journal={Advances in neural information processing systems},
  volume={35},
  pages={26565--26577},
  year={2022}
}

@inproceedings{esser2021taming,
  title={Taming transformers for high-resolution image synthesis},
  author={Esser, Patrick and Rombach, Robin and Ommer, Bjorn},
  booktitle={Proceedings of the IEEE/CVF conference on computer vision and pattern recognition},
  pages={12873--12883},
  year={2021}
}

@inproceedings{ramesh2021zero,
  title={Zero-shot text-to-image generation},
  author={Ramesh, Aditya and Pavlov, Mikhail and Goh, Gabriel and Gray, Scott and Voss, Chelsea and Radford, Alec and Chen, Mark and Sutskever, Ilya},
  booktitle={International conference on machine learning},
  pages={8821--8831},
  year={2021},
  organization={Pmlr}
}

@article{luong2015effective,
  title={Effective approaches to attention-based neural machine translation},
  author={Luong, Minh-Thang and Pham, Hieu and Manning, Christopher D},
  journal={arXiv preprint arXiv:1508.04025},
  year={2015}
}

@inproceedings{gu2022vector,
  title={Vector quantized diffusion model for text-to-image synthesis},
  author={Gu, Shuyang and Chen, Dong and Bao, Jianmin and Wen, Fang and Zhang, Bo and Chen, Dongdong and Yuan, Lu and Guo, Baining},
  booktitle={Proceedings of the IEEE/CVF conference on computer vision and pattern recognition},
  pages={10696--10706},
  year={2022}
}

@article{stiennon2020learning,
  title={Learning to summarize with human feedback},
  author={Stiennon, Nisan and Ouyang, Long and Wu, Jeffrey and Ziegler, Daniel and Lowe, Ryan and Voss, Chelsea and Radford, Alec and Amodei, Dario and Christiano, Paul F},
  journal={Advances in neural information processing systems},
  volume={33},
  pages={3008--3021},
  year={2020}
}

@article{ouyang2022training,
  title={Training language models to follow instructions with human feedback},
  author={Ouyang, Long and Wu, Jeffrey and Jiang, Xu and Almeida, Diogo and Wainwright, Carroll and Mishkin, Pamela and Zhang, Chong and Agarwal, Sandhini and Slama, Katarina and Ray, Alex and others},
  journal={Advances in neural information processing systems},
  volume={35},
  pages={27730--27744},
  year={2022}
}

@article{black2023training,
  title={Training diffusion models with reinforcement learning},
  author={Black, Kevin and Janner, Michael and Du, Yilun and Kostrikov, Ilya and Levine, Sergey},
  journal={arXiv preprint arXiv:2305.13301},
  year={2023}
}

@article{berthelot2023tract,
  title={Tract: Denoising diffusion models with transitive closure time-distillation},
  author={Berthelot, David and Autef, Arnaud and Lin, Jierui and Yap, Dian Ang and Zhai, Shuangfei and Hu, Siyuan and Zheng, Daniel and Talbott, Walter and Gu, Eric},
  journal={arXiv preprint arXiv:2303.04248},
  year={2023}
}

@article{li2024reward,
  title={Reward guided latent consistency distillation},
  author={Li, Jiachen and Feng, Weixi and Chen, Wenhu and Wang, William Yang},
  journal={arXiv preprint arXiv:2403.11027},
  year={2024}
}

@article{zheng2024trajectory,
  title={Trajectory Consistency Distillation},
  author={Zheng, Jianbin and Hu, Minghui and Fan, Zhongyi and Wang, Chaoyue and Ding, Changxing and Tao, Dacheng and Cham, Tat-Jen},
  journal={arXiv preprint arXiv:2402.19159},
  year={2024}
}

@inproceedings{xu2024ufogen,
  title={Ufogen: You forward once large scale text-to-image generation via diffusion gans},
  author={Xu, Yanwu and Zhao, Yang and Xiao, Zhisheng and Hou, Tingbo},
  booktitle={Proceedings of the IEEE/CVF Conference on Computer Vision and Pattern Recognition},
  pages={8196--8206},
  year={2024}
}

@article{sauer2023adversarial,
  title={Adversarial diffusion distillation},
  author={Sauer, Axel and Lorenz, Dominik and Blattmann, Andreas and Rombach, Robin},
  journal={arXiv preprint arXiv:2311.17042},
  year={2023}
}

@article{meng2022distillation,
  title={On distillation of guided diffusion models},
  author={Meng, Chenlin and Gao, Ruiqi and Kingma, Diederik P and Ermon, Stefano and Ho, Jonathan and Salimans, Tim},
  journal={arXiv preprint arXiv:2210.03142},
  year={2022}
}

@article{geng2024consistency,
  title={Consistency Models Made Easy},
  author={Geng, Zhengyang and Pokle, Ashwini and Luo, William and Lin, Justin and Kolter, J Zico},
  journal={arXiv preprint arXiv:2406.14548},
  year={2024}
}

@article{salimans2024multistep,
  title={Multistep Distillation of Diffusion Models via Moment Matching},
  author={Salimans, Tim and Mensink, Thomas and Heek, Jonathan and Hoogeboom, Emiel},
  journal={arXiv preprint arXiv:2406.04103},
  year={2024}
}

@article{xie2024distillation,
  title={EM Distillation for One-step Diffusion Models},
  author={Xie, Sirui and Xiao, Zhisheng and Kingma, Diederik P and Hou, Tingbo and Wu, Ying Nian and Murphy, Kevin Patrick and Salimans, Tim and Poole, Ben and Gao, Ruiqi},
  journal={arXiv preprint arXiv:2405.16852},
  year={2024}
}

@article{heek2024multistep,
  title={Multistep consistency models},
  author={Heek, Jonathan and Hoogeboom, Emiel and Salimans, Tim},
  journal={arXiv preprint arXiv:2403.06807},
  year={2024}
}

@article{zhou2024long,
  title={Long and Short Guidance in Score identity Distillation for One-Step Text-to-Image Generation},
  author={Zhou, Mingyuan and Wang, Zhendong and Zheng, Huangjie and Huang, Hai},
  journal={arXiv preprint arXiv:2406.01561},
  year={2024}
}

@article{luo2024diff,
  title={Diff-instruct: A universal approach for transferring knowledge from pre-trained diffusion models},
  author={Luo, Weijian and Hu, Tianyang and Zhang, Shifeng and Sun, Jiacheng and Li, Zhenguo and Zhang, Zhihua},
  journal={Advances in Neural Information Processing Systems},
  volume={36},
  year={2024}
}

@article{yin2023one,
  title={One-step Diffusion with Distribution Matching Distillation},
  author={Yin, Tianwei and Gharbi, Micha{\"e}l and Zhang, Richard and Shechtman, Eli and Durand, Fredo and Freeman, William T and Park, Taesung},
  journal={arXiv preprint arXiv:2311.18828},
  year={2023}
}

@article{kim2023consistency,
  title={Consistency trajectory models: Learning probability flow ode trajectory of diffusion},
  author={Kim, Dongjun and Lai, Chieh-Hsin and Liao, Wei-Hsiang and Murata, Naoki and Takida, Yuhta and Uesaka, Toshimitsu and He, Yutong and Mitsufuji, Yuki and Ermon, Stefano},
  journal={arXiv preprint arXiv:2310.02279},
  year={2023}
}

@inproceedings{songimproved,
  title={Improved Techniques for Training Consistency Models},
  author={Song, Yang and Dhariwal, Prafulla},
  booktitle={The Twelfth International Conference on Learning Representations},
  year=2024,
}

@article{van2016conditional,
  title={Conditional image generation with pixelcnn decoders},
  author={Van den Oord, Aaron and Kalchbrenner, Nal and Espeholt, Lasse and Vinyals, Oriol and Graves, Alex and others},
  journal={Advances in neural information processing systems},
  volume={29},
  year={2016}
}

@inproceedings{reed2017parallel,
  title={Parallel multiscale autoregressive density estimation},
  author={Reed, Scott and Oord, A{\"a}ron and Kalchbrenner, Nal and Colmenarejo, Sergio G{\'o}mez and Wang, Ziyu and Chen, Yutian and Belov, Dan and Freitas, Nando},
  booktitle={International conference on machine learning},
  pages={2912--2921},
  year={2017},
  organization={PMLR}
}

@article{reed2017few,
  title={Few-shot autoregressive density estimation: Towards learning to learn distributions},
  author={Reed, Scott and Chen, Yutian and Paine, Thomas and Oord, A{\"a}ron van den and Eslami, SM and Rezende, Danilo and Vinyals, Oriol and de Freitas, Nando},
  journal={arXiv preprint arXiv:1710.10304},
  year={2017}
}

@article{tian2024visual,
  title={Visual autoregressive modeling: Scalable image generation via next-scale prediction},
  author={Tian, Keyu and Jiang, Yi and Yuan, Zehuan and Peng, Bingyue and Wang, Liwei},
  journal={Advances in neural information processing systems},
  volume={37},
  pages={84839--84865},
  year={2024}
}

@article{luo2025reward,
  title={Reward-Instruct: A Reward-Centric Approach to Fast Photo-Realistic Image Generation},
  author={Luo, Yihong and Hu, Tianyang and Luo, Weijian and Kawaguchi, Kenji and Tang, Jing},
  journal={arXiv preprint arXiv:2503.13070},
  year={2025}
}

@article{wang2025uni,
  title={Uni-Instruct: One-step Diffusion Model through Unified Diffusion Divergence Instruction},
  author={Wang, Yifei and Bai, Weimin and Zhang, Colin and Zhang, Debing and Luo, Weijian and Sun, He},
  journal={arXiv preprint arXiv:2505.20755},
  year={2025}
}

@article{luo2023comprehensive,
  title={A Comprehensive Survey on Knowledge Distillation of Diffusion Models},
  author={Luo, Weijian},
  journal={arXiv preprint arXiv:2304.04262},
  year={2023}
}

@article{luo2024diffpp,
  title={Diff-instruct++: Training one-step text-to-image generator model to align with human preferences},
  author={Luo, Weijian},
  journal={arXiv preprint arXiv:2410.18881},
  year={2024}
}

@article{luo2024david,
  title={David and Goliath: Small One-step Model Beats Large Diffusion with Score Post-training},
  author={Luo, Weijian and Zhang, Colin and Zhang, Debing and Geng, Zhengyang},
  journal={arXiv preprint arXiv:2410.20898},
  year={2024}
}

@article{luo2024diffstar,
  title={Diff-instruct*: Towards human-preferred one-step text-to-image generative models},
  author={Luo, Weijian and Zhang, Colin and Zhang, Debing and Geng, Zhengyang},
  journal={arXiv e-prints},
  pages={arXiv--2410},
  year={2024}
}

@inproceedings{hong2024margin,
  title={Margin-aware preference optimization for aligning diffusion models without reference},
  author={Hong, Jiwoo and Paul, Sayak and Lee, Noah and Rasul, Kashif and Thorne, James and Jeong, Jongheon},
  booktitle={First Workshop on Scalable Optimization for Efficient and Adaptive Foundation Models},
  year={2024}
}

@inproceedings{yang2024using,
  title={Using human feedback to fine-tune diffusion models without any reward model},
  author={Yang, Kai and Tao, Jian and Lyu, Jiafei and Ge, Chunjiang and Chen, Jiaxin and Shen, Weihan and Zhu, Xiaolong and Li, Xiu},
  booktitle={Proceedings of the IEEE/CVF Conference on Computer Vision and Pattern Recognition},
  pages={8941--8951},
  year={2024}
}

@inproceedings{wallace2024diffusion,
  title={Diffusion model alignment using direct preference optimization},
  author={Wallace, Bram and Dang, Meihua and Rafailov, Rafael and Zhou, Linqi and Lou, Aaron and Purushwalkam, Senthil and Ermon, Stefano and Xiong, Caiming and Joty, Shafiq and Naik, Nikhil},
  booktitle={Proceedings of the IEEE/CVF Conference on Computer Vision and Pattern Recognition},
  pages={8228--8238},
  year={2024}
}

@article{fan2024reinforcement,
  title={Reinforcement learning for fine-tuning text-to-image diffusion models},
  author={Fan, Ying and Watkins, Olivia and Du, Yuqing and Liu, Hao and Ryu, Moonkyung and Boutilier, Craig and Abbeel, Pieter and Ghavamzadeh, Mohammad and Lee, Kangwook and Lee, Kimin},
  journal={Advances in Neural Information Processing Systems},
  volume={36},
  year={2024}
}

@article{lee2023aligning,
  title={Aligning text-to-image models using human feedback},
  author={Lee, Kimin and Liu, Hao and Ryu, Moonkyung and Watkins, Olivia and Du, Yuqing and Boutilier, Craig and Abbeel, Pieter and Ghavamzadeh, Mohammad and Gu, Shixiang Shane},
  journal={arXiv preprint arXiv:2302.12192},
  year={2023}
}

@article{clark2023directly,
  title={Directly fine-tuning diffusion models on differentiable rewards},
  author={Clark, Kevin and Vicol, Paul and Swersky, Kevin and Fleet, David J},
  journal={arXiv preprint arXiv:2309.17400},
  year={2023}
}

@article{prabhudesai2023aligning,
  title={Aligning text-to-image diffusion models with reward backpropagation},
  author={Prabhudesai, Mihir and Goyal, Anirudh and Pathak, Deepak and Fragkiadaki, Katerina},
  journal={arXiv preprint arXiv:2310.03739},
  year={2023}
}

@article{dai2023emu,
  title={Emu: Enhancing image generation models using photogenic needles in a haystack},
  author={Dai, Xiaoliang and Hou, Ji and Ma, Chih-Yao and Tsai, Sam and Wang, Jialiang and Wang, Rui and Zhang, Peizhao and Vandenhende, Simon and Wang, Xiaofang and Dubey, Abhimanyu and others},
  journal={arXiv preprint arXiv:2309.15807},
  year={2023}
}

@article{podell2023sdxl,
  title={Sdxl: Improving latent diffusion models for high-resolution image synthesis},
  author={Podell, Dustin and English, Zion and Lacey, Kyle and Blattmann, Andreas and Dockhorn, Tim and M{\"u}ller, Jonas and Penna, Joe and Rombach, Robin},
  journal={arXiv preprint arXiv:2307.01952},
  year={2023}
}

@article{vincent2011connection,
  title={A connection between score matching and denoising autoencoders},
  author={Vincent, Pascal},
  journal={Neural computation},
  volume={23},
  number={7},
  pages={1661--1674},
  year={2011},
  publisher={MIT Press}
}

@inproceedings{wu2025dc,
  title={Dc-ar: Efficient masked autoregressive image generation with deep compression hybrid tokenizer},
  author={Wu et.al},
  booktitle={ICCV},
  pages={18034--18045},
  year={2025}
}

@article{sun2023journeydb,
  title={Journeydb: A benchmark for generative image understanding},
  author={Sun, Keqiang and Pan, Junting and Ge, Yuying and Li, Hao and Duan, Haodong and Wu, Xiaoshi and Zhang, Renrui and Zhou, Aojun and Qin, Zipeng and Wang, Yi and others},
  journal={Advances in neural information processing systems},
  volume={36},
  pages={49659--49678},
  year={2023}
}
}

\clearpage
\setcounter{page}{1}
\setcounter{section}{0}
\setcounter{figure}{0}
\setcounter{table}{0}
\maketitlesupplementary

\section{Theory Part}
\label{sec:Proof}
In this section we present a detailed proof of the Gradient Equivalent
Theorem. The proof of the Gradient Equivalent Theorem is based on the so-called \textbf{Score-projection identity} which was first found in ~\cite{vincent2011connection} to bridge denoising score matching and denoising auto-encoders. Later the identity is applied for deriving distillation methods based on Fisher divergences ~\cite{zhou2024score}. We appreciate the efforts of Zhou et al. ~\cite{zhou2024score} and re-write the score-projection identity here without proof. Readers can check Zhou et al. ~\cite{zhou2024score} for a complete proof of score-projection identity.

\paragraph{Score-projection identity.} 
Let $u(\cdot,\theta)$ be a vector-valued function, using the notations of the Gradient Equivalent Theorem, under mild conditions, the identity holds:
\begin{align*}
&\mathbb{E}_{\substack{ 
        \boldsymbol{x}_0 \sim q_{\theta,0} \\
        \boldsymbol{x}_t | \boldsymbol{x}_0 \sim p_t(\boldsymbol{x}_t | \boldsymbol{x}_0)
        }} 
\Big[f(\boldsymbol{x}_t, \boldsymbol{x}_0, \theta)\Big]
=0, \quad \forall \theta \tag{A.1} \\
&f(\boldsymbol{x}_t, \boldsymbol{x}_0, \theta)= u(\boldsymbol{x}_t,\theta)^{T} \big\{ s_{q_{\theta,t}}(\boldsymbol{x}_t,c)
- \nabla_{\boldsymbol{x}_t}\log p_{t}(\boldsymbol{x}_t | \boldsymbol{x}_0, c)\big\}.
\end{align*}
Next, we turn to prove the Gradient Equivalent Theorem.

\paragraph{Proof of Gradient Equivalent Theorem.}
We start by applying the chain rule for the total derivative with respect to $\theta$. The function $d(\cdot)$ depends on $\theta$ both directly through the score function $s_{q_{\theta,t}}$ and indirectly through the distribution $\boldsymbol{x}_t \sim q_{\theta,t}$ (as $\boldsymbol{x}_t$ depends on $\boldsymbol{x}_0 \sim q_{\theta, 0}$). This gives two terms:
\begin{align*}
& \mathbb{E}_{\boldsymbol{x}_t \sim q_{\theta,t}}  \frac{\partial}{\partial\theta} d \left( s_{q_{\theta,t}}(\boldsymbol{x}_t, c) - s_{p_t}(\boldsymbol{x}_t, c) \right) \\ 
    &=\mathbb{E}_{\boldsymbol{x}_t \sim q_{\theta,t}} \Biggl[ d' \left( s_{q_{\theta,t}}(\boldsymbol{x}_t, c) - s_{p_t}(\boldsymbol{x}_t, c) \right)^T \frac{\partial}{\partial\theta} s_{q_{\theta,t}}(\boldsymbol{x}_t, c) \\ 
    &+ \frac{\partial}{\partial \boldsymbol{x}_t }d(s_{q_{\theta,t}}(\boldsymbol{x}_t, c) - s_{p_t}(\boldsymbol{x}_t, c)) \frac{\partial \boldsymbol{x}_t }{\partial \theta } \Biggr] \tag{A.2}
\end{align*}

We differentiate (A.1) on both sides w.r.t $\theta$ to achieve the equivalent of the first term in (A.2). Since the expectation is zero for all $\theta$, its derivative is also zero. We apply the total derivative (multivariate chain rule), noting that $f$ depends on $\theta$ directly, as well as indirectly through $\boldsymbol{x}_0$ and $\boldsymbol{x}_t$ (both depend on $\theta$):
\begin{align*}
0
&= \frac{\partial}{\partial\theta}\,
\mathbb{E}_{\substack{ 
        \boldsymbol{x}_0 \sim q_{\theta,0} \\
        \boldsymbol{x}_t | \boldsymbol{x}_0 \sim p_t(\boldsymbol{x}_t | \boldsymbol{x}_0)
        }}
\Big[ f(\boldsymbol{x}_t, \boldsymbol{x}_0, \theta)\Big]
\tag{A.3}\\
&= \mathbb{E}
\Big[ \frac{\partial f}{\partial \boldsymbol{x}_0}\frac{\partial \boldsymbol{x}_0}{\partial\theta} +\frac{\partial f}{\partial \boldsymbol{x}_t}\frac{\partial \boldsymbol{x}_t}{\partial\theta} +\frac{\partial f}{\partial \theta}
\Big]\tag{A.4}\\
&= \mathbb{E}
\Big[\frac{\partial}{\partial \boldsymbol{x}_0} \Big\{ u(x_t, \theta)^T \{-\nabla_{x_t} \log p_t(x_t|x_0, c) \} \Big\} \frac{\partial x_0}{\partial \theta} \tag{A.5}\\
&+ \frac{\partial}{\partial \boldsymbol{x}_t} \Big\{ u(\boldsymbol{x}_t, \theta)^T \{ s_{q_{\theta,t}}(x_t, c) - \nabla_{\boldsymbol{x}_t} \log p_t(\boldsymbol{x}_t|x_0, c) \} \Big\} \frac{\partial \boldsymbol{x}_t}{\partial \theta}\\
&+ \frac{\partial}{\partial \theta} u(\boldsymbol{x}_t, \theta)^T s_{q_{\theta,t}}(\boldsymbol{x}_t, c) + u(\boldsymbol{x}_t, \theta)^T \frac{\partial}{\partial \theta} \{ s_{q_{\theta,t}}(\boldsymbol{x}_t, c) \}
\Big] \\
&= \mathbb{E} \Big[ 
\frac{\partial}{\partial \theta}  \Big\{ u(x_t, \theta)^T \{ s_{q_{\text{sg}[\theta],t}}(x_t, c) - \nabla_{x_t} \log p_t(x_t|x_0, c) \} \Big\} \\
&+ u(\boldsymbol{x}_t, \theta)^T \frac{\partial}{\partial \theta} \{ s_{q_{\theta,t}}(\boldsymbol{x}_t, c)
\Big]. \tag{A.6} \\
\end{align*}

By rearranging the terms in (A.4) and (A.5), and by applying the stop-gradient operator $\text{sg}$ to $s_{q_{\theta,t}}$ within the derivative terms, the expression can be consolidated into (A.6). We then rewrite (A.6) as follows:

\begin{align*}
&\mathbb{E}_{x_t \sim q_{\theta,t}} u(\boldsymbol{x}_t, \theta)^T \frac{\partial}{\partial \theta} \{ s_{q_{\theta,t}}(\boldsymbol{x}_t, c) \} \\
&= -\frac{\partial}{\partial \theta} 
\mathbb{E}_{\substack{ 
        \boldsymbol{x}_0 \sim q_{\theta,0} \\
        \boldsymbol{x}_t | \boldsymbol{x}_0 \sim p_t(\boldsymbol{x}_t | \boldsymbol{x}_0)
        }}
\Big\{ u(x_t, \theta)^T \\
&\{ s_{q_{\text{sg}[\theta],t}}(x_t, c) - \nabla_{x_t} \log p_t(x_t|x_0, c) \} \Big\} \tag{A.7}
\end{align*}

Let $ u(\boldsymbol{x}_t, \theta) = d' \left( s_{q_{\text{sg}[\theta],t}}(\boldsymbol{x}_t, c) - s_{p_t}(\boldsymbol{x}_t, c) \right)$ and substitute this specific function $u$ into the identity (A.7):
\begin{align*}
&\mathbb{E}_{x_t \sim q_{\theta,t}} \Big[ d' \left( s_{q_{\text{sg}[\theta],t}}(\boldsymbol{x}_t, c) - s_{p_t}(\boldsymbol{x}_t, c) \right)^T \frac{\partial}{\partial \theta} \{ s_{q_{\theta,t}}(\boldsymbol{x}_t, c) \} \Big] \tag{A.8}\\
&= \mathbb{E}_{x_t \sim q_{\theta,t}} \Big[ d' \left( s_{q_{\theta,t}}(\boldsymbol{x}_t, c) - s_{p_t}(\boldsymbol{x}_t, c) \right)^T \frac{\partial}{\partial \theta} \{ s_{q_{\theta,t}}(\boldsymbol{x}_t, c) \} \Big] \tag{A.9}\\
&= -\frac{\partial}{\partial \theta} 
\mathbb{E}_{\substack{ 
        \boldsymbol{x}_0 \sim q_{\theta,0} \\
        \boldsymbol{x}_t | \boldsymbol{x}_0 \sim p_t(\boldsymbol{x}_t | \boldsymbol{x}_0)
        }}
\Big\{ d' \left( s_{q_{\text{sg}[\theta],t}}(\boldsymbol{x}_t, c) - s_{p_t}(\boldsymbol{x}_t, c) \right)^T \\
&\{ s_{q_{\text{sg}[\theta],t}}(x_t, c) - \nabla_{x_t} \log p_t(x_t|x_0, c) \} \Big\} \tag{A.10}
\end{align*}

Since $\theta$ does not participate in the derivative of $d(\cdot)$, equation (A.8) is equivalent to (A.9). Moreover, $\theta$ does not appear in the differentiation with respect to $\boldsymbol{x}_t$:
\begin{align}
&\mathbb{E}_{x_t \sim q_{\theta,t}} \Big[ \frac{\partial}{\partial \boldsymbol{x}_t }d(s_{q_{\theta,t}}(\boldsymbol{x}_t, c) - s_{p_t}(\boldsymbol{x}_t, c)) \frac{\partial \boldsymbol{x}_t }{\partial \theta }\Big] = \nonumber \\
&-\frac{\partial}{\partial \theta} 
\mathbb{E}_{\substack{ 
        \boldsymbol{x}_0 \sim q_{\theta,0} \\
        \boldsymbol{x}_t | \boldsymbol{x}_0 \sim p_t(\boldsymbol{x}_t | \boldsymbol{x}_0)
        }} \Big[ d \left( s_{q_{\text{sg}[\theta],t}}(\boldsymbol{x}_t, c) - s_{p_t}(\boldsymbol{x}_t, c) \right) \Big] \tag{A.11}
\end{align}

By combining (A.10) and (A.11) into (A.2), we get exactly the \textbf{Gradient Equivalent Theorem} stated in the text.
This completes the proof.

\section{Implementation Details}
\label{sec:implementation}

In this section, we provide detailed hyperparameter settings and the algorithmic procedure for MARVAL. Our framework consists of two distinct stages: the Guided Score Implicit Matching (GSIM) distillation stage and the Reinforcement Learning (RL) refinement stage.


\subsection{Hyperparameter Settings}

\paragraph{Model Architectures.}
The architecture of our one-step generator follows the same configuration of the original MAR models~\cite{zhou2024score}.

\begin{table}[ht]
    \centering
    \begin{tabular}{l|c}
        \toprule
        \textbf{Hyperparameter} & \textbf{Stage 1: GSIM Distillation} \\
        \midrule
        \textbf{GPUs} & 8 $\times$ NVIDIA A100 \\
        \textbf{Training Epochs} & 30  \\
        \textbf{Approx. Training Time} & $\sim$ 3 days  \\
        \textbf{Teacher Steps ($N_{diff}$)} & 1000 \\
        \textbf{Student Steps ($N_{diff}$)} & 1 \\
        \textbf{CFG Scale ($w$)} & 1.2 \\
        \textbf{Loss Function} & Pseudo-Huber ($r=10^{-5}$) \\
        \textbf{Student model Lr} & 5e-6 \\
        \textbf{Auxiliary model Lr} & 5e-6 \\
        \textbf{EMA momentum} & 0.9999 \\
        \textbf{batch size per GPU} & 64 \\
        \bottomrule
    \end{tabular}
    \caption{Training Settings for MARVAL Distillation stage.}
    \label{tab:distill_para}
\end{table}

\begin{table}[ht]
    \centering
    \begin{tabular}{l|c}
        \toprule
        \textbf{Hyperparameter} & \textbf{Stage 2: RL Refinement} \\
        \midrule
        \textbf{GPUs} & 8 $\times$ NVIDIA A100 \\
        \textbf{Training Epochs} & 5  \\
        \textbf{Approx. Training Time} & $\sim$ 2 days  \\
        \textbf{AR Loops ($N_{AR}$)} & 64 \\
        \textbf{Student Steps ($N_{diff}$)} & 1 \\
        \textbf{Reward Model} & Pickscore \\
        \textbf{Student model Lr} & 5e-6 \\
        \textbf{EMA momentum} & 0.9999 \\
        \textbf{batch size per GPU} & 2 \\
        \bottomrule
    \end{tabular}
    \caption{Training Settings for RL Refinement stage.}
    \label{tab:rl_para}
\end{table}

\paragraph{Training Configurations.}
The training is conducted on NVIDIA A100 GPUs. We utilize the AdamW optimizer for both stages. The specific hyperparameters for the Distillation and RL stages are provided in Table~\ref{tab:distill_para} and Table~\ref{tab:rl_para}.

During the distillation stage, we set the teacher's full diffusion steps to $N_{diff}=1000$ and the one-step generator predicts the noise from the fix step 400. Based on our ablation study, we set the Classifier-Free Guidance (CFG) scale $w=1.2$ for the teacher score, which provides the optimal trade-off between fidelity (FID) and fidelity. For the loss function, we use the Pseudo-Huber distance with $r=1e-5$ to ensure numerical stability.

During the RL refinement stage, we use PickScore~\cite{kirstain2023pick} as the reward model. We generate samples with AR loops($N_{AR}$)=64 to calculate rewards, ensuring the optimization aligns with the final inference quality and requires less memory costs. Our best results in FID and IS are tested when inferring with $N_{AR}=128$.

\section{Additional Results}

This section provides additional qualitative results for the MARVAL-RL-L and MARVAL-RL-H models, as well as extended text-to-image generation samples from our 1-step+RL DC-AR model. The MARVAL results were not included in the main paper for the following reasons:
\begin{itemize}
\item First, due to space limitations, we primarily presented results using the MARVAL-RL-B model. 
\item Second, the qualitative visualizations in the main experiments adopt MARVAL-RL-B because it already delivers strong visual quality while maintaining the fastest inference speed among all model sizes. 
\item Third, the MARVAL-L and MARVAL-H models achieve relatively strong performance even before RL fine-tuning, making the improvements brought by RL most pronounced and interpretable on MARVAL-B.
\end{itemize}

\begin{figure*}[t]
  \centering
   \includegraphics[width=1.0 \linewidth]{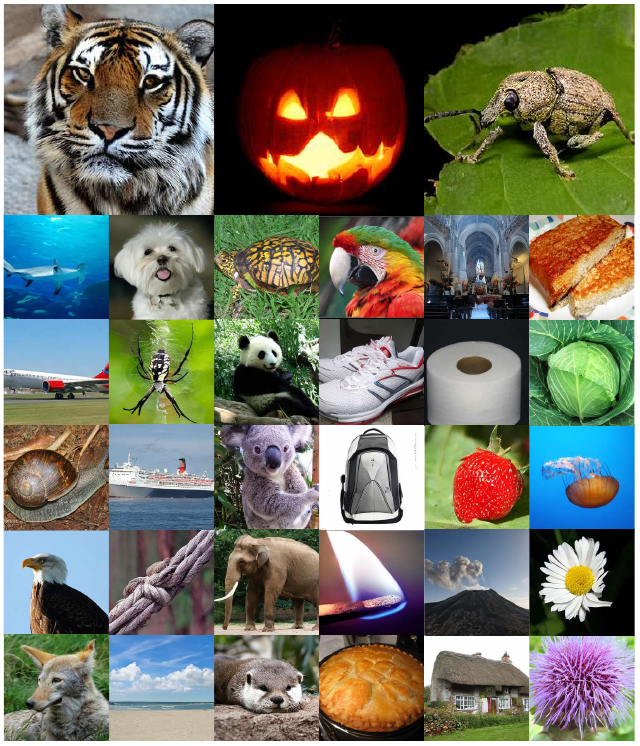} 
    \caption{Qualitative results of MARVAL-RL-L.}
    \label{fig:large}
\end{figure*}

\begin{figure*}[t]
  \centering
   \includegraphics[width=1.0 \linewidth]{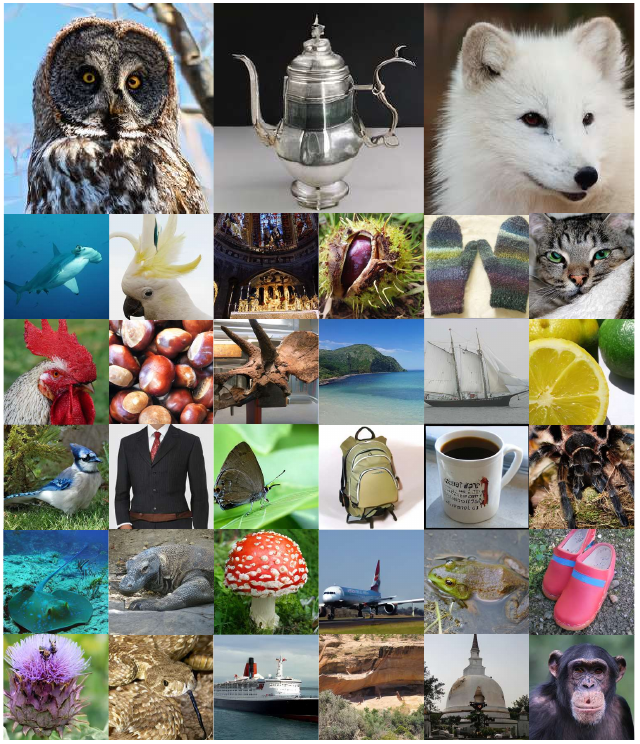} 
    \caption{Qualitative results of MARVAL-RL-H.}
    \label{fig:huge}
\end{figure*}

For completeness, we include the MARVAL-RL-L and MARVAL-RL-H visualizations in Fig.~\ref{fig:large} and Fig.~\ref{fig:huge}, respectively. Compared with MARVAL-RL-B and MARVAL-RL-L, the MARVAL-RL-H samples exhibit significantly richer fine-grained textures, particularly on man-made objects such as buildings, tools, and clothes. This trend provides additional evidence that our method scales effectively: as model capacity increases, the RL fine-tuning yields increasingly detailed and visually faithful results.

\begin{figure*}[htbp]
    \centering
    \begin{minipage}[t]{0.24\linewidth}
        \centering
        \includegraphics[width=\linewidth]{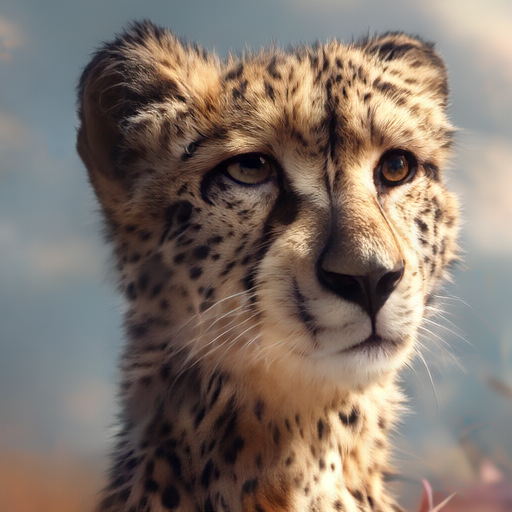}
        \vspace{0.5mm}
        \scriptsize Cheetah, pastel, cinematic lighting, full body view, stylize, detailed, v4.
    \end{minipage}\hfill
    \begin{minipage}[t]{0.24\linewidth}
        \centering
        \includegraphics[width=\linewidth]{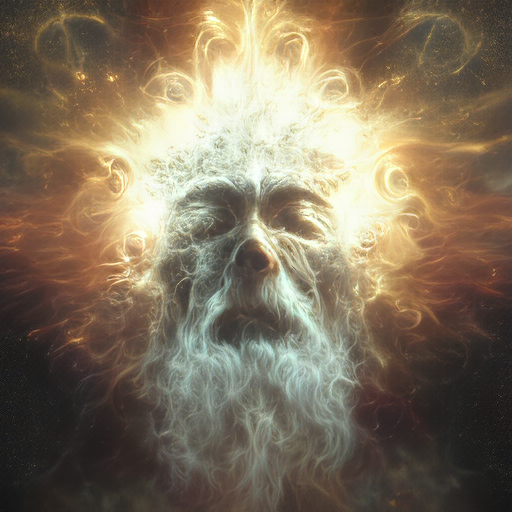}
        \vspace{0.5mm}
        \scriptsize The face of god, light, holy.
    \end{minipage}\hfill
    \begin{minipage}[t]{0.24\linewidth}
        \centering
        \includegraphics[width=\linewidth]{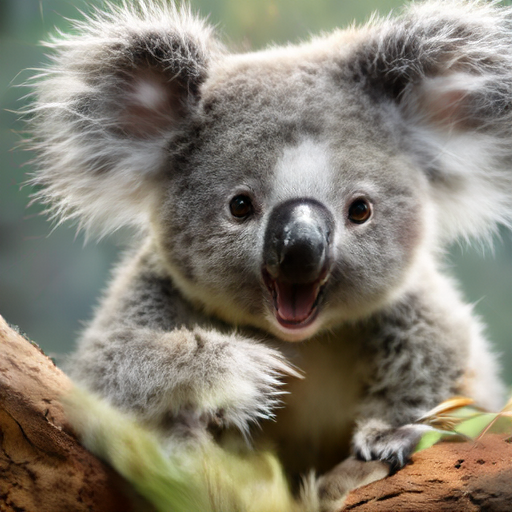}
        \vspace{0.5mm}
        \scriptsize Fluffy koala bear happy.
    \end{minipage}\hfill
    \begin{minipage}[t]{0.24\linewidth}
        \centering
        \includegraphics[width=\linewidth]{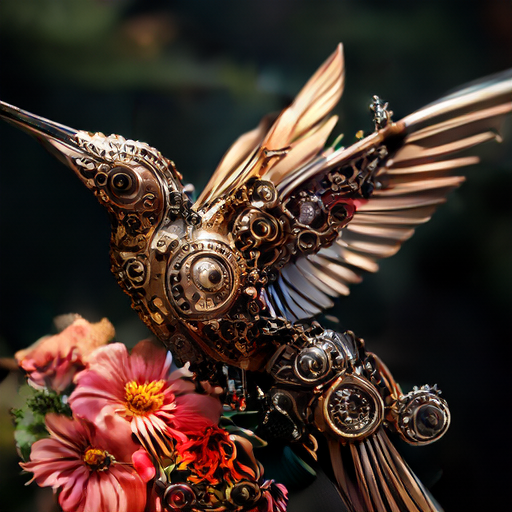}
        \vspace{0.5mm}
        \scriptsize Mechanical hummingbird made of brass and copper, hovering near flowers, intricate gears, steampunk style.
    \end{minipage}

    \begin{minipage}[t]{0.24\linewidth}
        \centering
        \includegraphics[width=\linewidth]{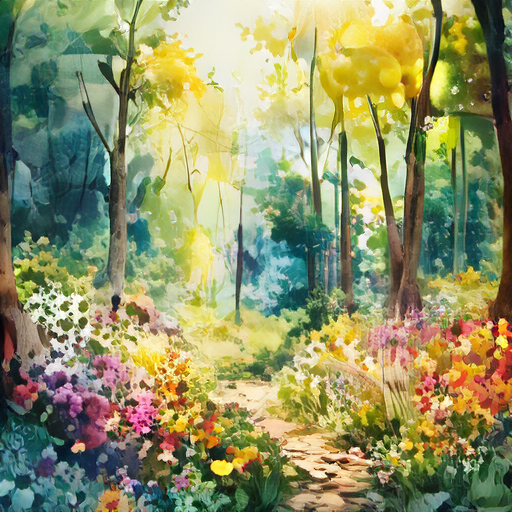}
        \vspace{0.5mm} 
        \scriptsize A watercolor of a beautiful forest with flowers, sunny, colorful.
    \end{minipage}\hfill
    \begin{minipage}[t]{0.24\linewidth}
        \centering
        \includegraphics[width=\linewidth]{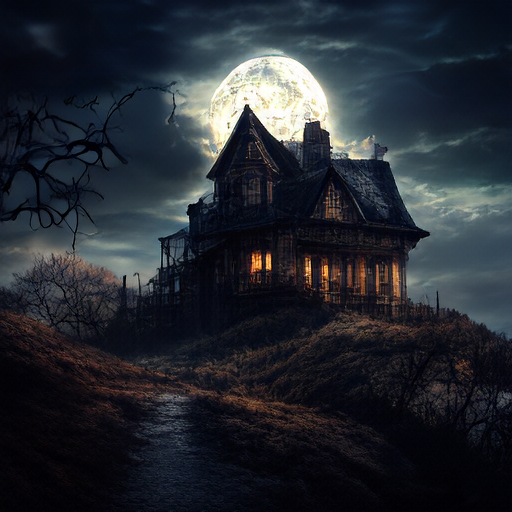}
        \vspace{0.5mm}
        \scriptsize A haunted house on a hill under a full moon.
    \end{minipage}\hfill
    \begin{minipage}[t]{0.24\linewidth}
        \centering
        \includegraphics[width=\linewidth]{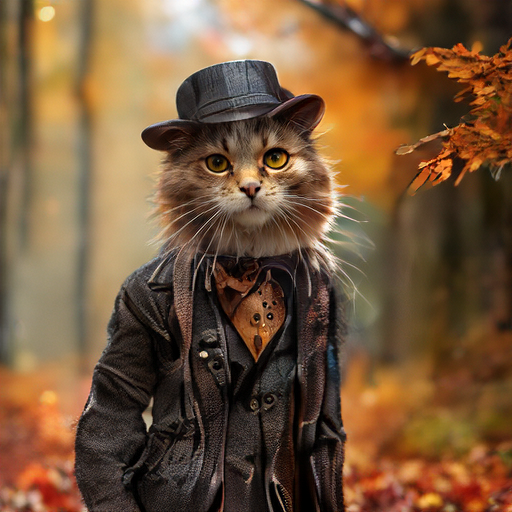}
        \vspace{0.5mm}
        \scriptsize Professional portrait of an anthropomorphic cat wearing gentleman hat and jacket walking in autumn forest.
    \end{minipage}\hfill
    \begin{minipage}[t]{0.24\linewidth}
        \centering
        \includegraphics[width=\linewidth]{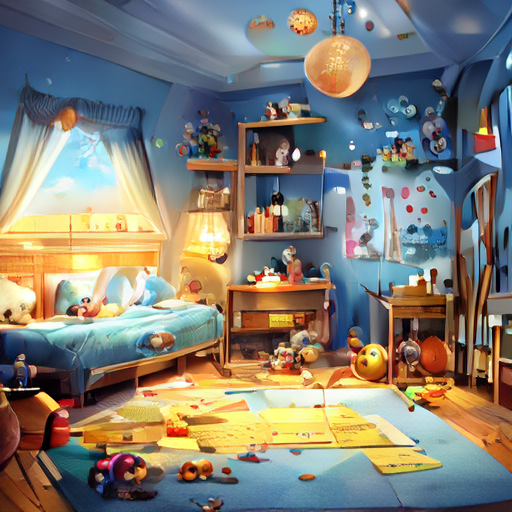}
        \vspace{0.5mm}
        \scriptsize Boy's room, children's room, comfortable room with toys, Freelancer game, drawing for a children's storybook, blue room, cartoon.
    \end{minipage}

    \begin{minipage}[t]{0.24\linewidth}
        \centering
        \includegraphics[width=\linewidth]{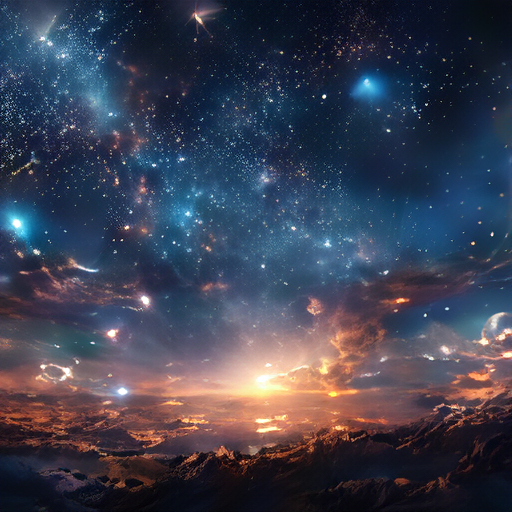}
        \vspace{0.5mm} 
        \scriptsize Starfield, deep sky, realistic, 8k
    \end{minipage}\hfill
    \begin{minipage}[t]{0.24\linewidth}
        \centering
        \includegraphics[width=\linewidth]{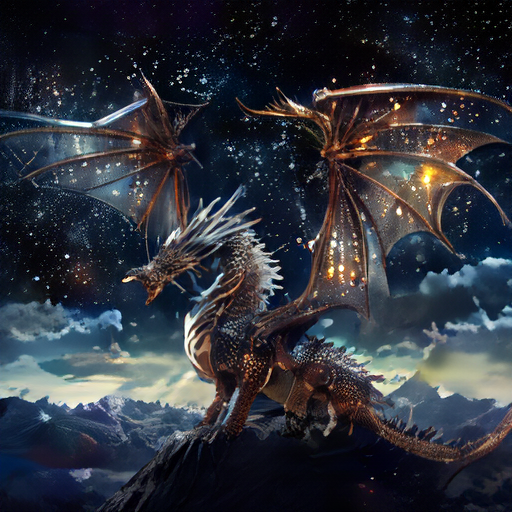}
        \vspace{0.5mm}
        \scriptsize Dragon made of constellation stars flying across night sky, over mountain landscape.
    \end{minipage}\hfill
    \begin{minipage}[t]{0.24\linewidth}
        \centering
        \includegraphics[width=\linewidth]{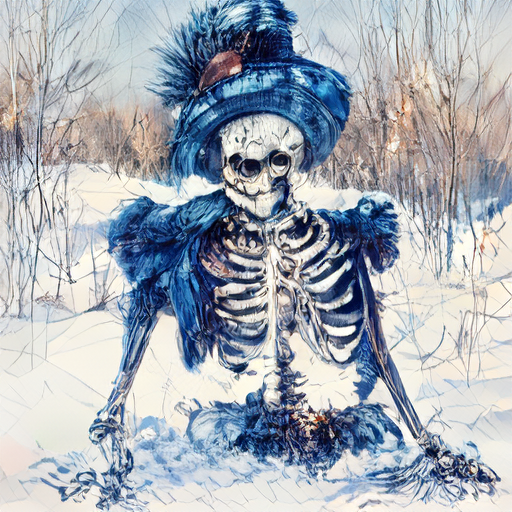}
        \vspace{0.5mm}
        \scriptsize Skeleton, in the snow, friendly, made with pen and ink and prismacolor shading, blue and white color palet.
    \end{minipage}\hfill
    \begin{minipage}[t]{0.24\linewidth}
        \centering
        \includegraphics[width=\linewidth]{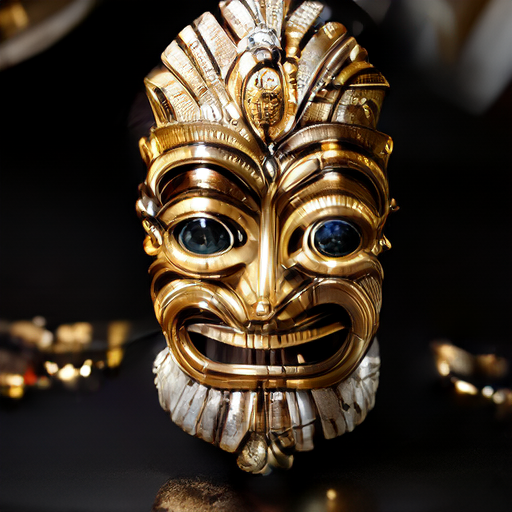}
        \vspace{0.5mm}
        \scriptsize An elegant gold and white tikka head jewelry.
    \end{minipage}

    \begin{minipage}[t]{0.24\linewidth}
        \centering
        \includegraphics[width=\linewidth]{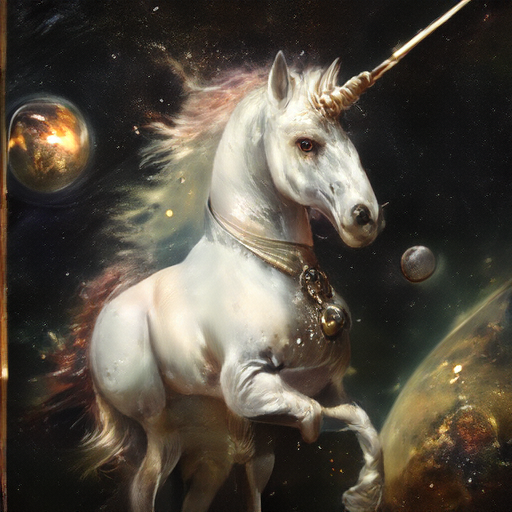}
        \vspace{0.5mm} 
        \scriptsize A framed renaissance painting of a unicorn on another planet in outer space. oil painting on canvass. rembrandt. 1647.
    \end{minipage}\hfill
    \begin{minipage}[t]{0.24\linewidth}
        \centering
        \includegraphics[width=\linewidth]{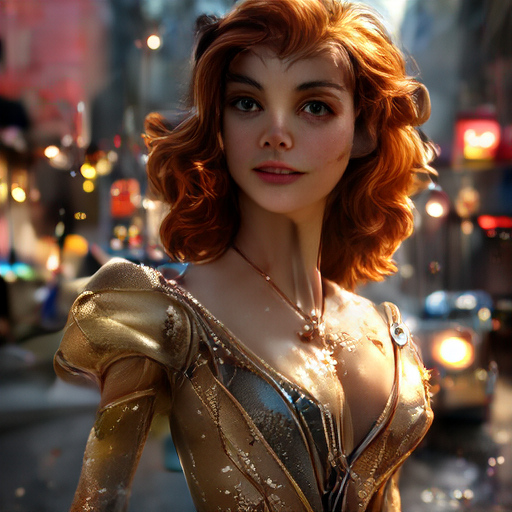}
        \vspace{0.5mm}
        \scriptsize Christina Hendricks as a Disney princess, Full Torso, standing on a city street, octane render, volumetric lighting.
    \end{minipage}\hfill
    \begin{minipage}[t]{0.24\linewidth}
        \centering
        \includegraphics[width=\linewidth]{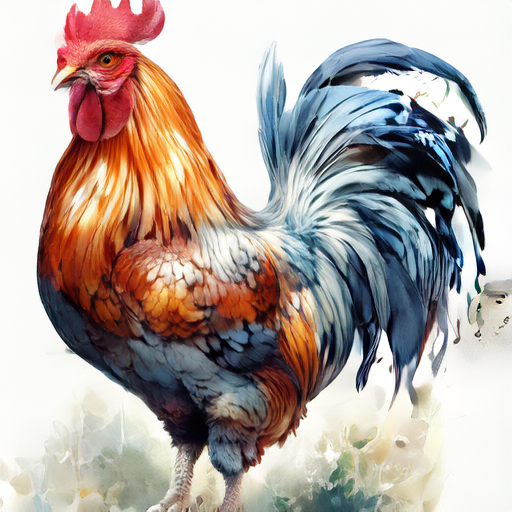}
        \vspace{0.5mm} 
        \scriptsize A painting of a rooster on a white background, a watercolor painting by Sun Kehong, shutterstock contest winner, cloisonnism, detailed painting, behance hd, photoillustration, 8K, UHD.
    \end{minipage}\hfill
    \begin{minipage}[t]{0.24\linewidth}
        \centering
        \includegraphics[width=\linewidth]{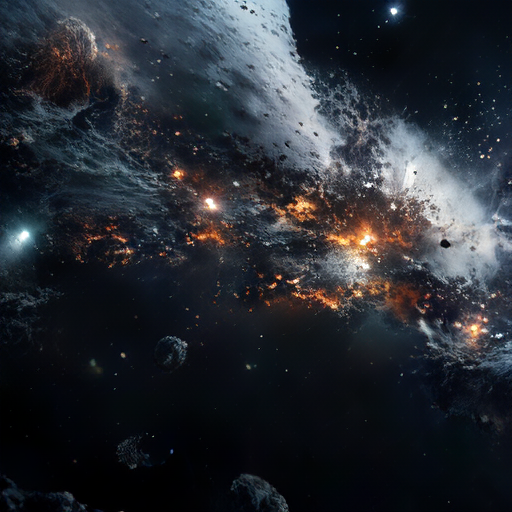}
        \vspace{0.5mm}
        \scriptsize Starfield, space, 8k, photo, realism, sharp photography, maximum detail, sharp focus, Intricate details, epic, wide shot, highly realistic, cinematic lighting, volumetric lighting, octane render.
    \end{minipage}

    \caption{More qualitative results of distilled+RL DC-AR text-to-image generation. Our method achieves high generation quality and strong text alignment, demonstrating robust adaptability to prompts of varying lengths.}
    \label{fig:dc-ar}
\end{figure*}

Furthermore, to supplement the text-to-image experiments presented in the main text, we provide a broader set of qualitative samples generated by our 1-step+RL DC-AR model in Fig.~\ref{fig:dc-ar}. 

\end{document}